\documentclass{article} % For LaTeX2e
\usepackage{iclr2026_conference,times}

\usepackage{amsmath,amsfonts,bm}

\def\eqref#1{equation~\ref{#1}}
\def\1{\bm{1}}

\DeclareMathAlphabet{\mathsfit}{\encodingdefault}{\sfdefault}{m}{sl}
\SetMathAlphabet{\mathsfit}{bold}{\encodingdefault}{\sfdefault}{bx}{n}

\usepackage{hyperref}
\usepackage{url}
\usepackage{placeins}

\usepackage{xcolor}         % colors
\usepackage{graphicx}
\usepackage{multirow}
 \usepackage{amsmath} 
  \usepackage{makecell} 
\usepackage{amsfonts}       % blackboard math symbols
\usepackage{booktabs}
\usepackage{natbib}     % for \shortcite (if needed)
\usepackage{wrapfig} % preamble에 추가

\newcommand{\proj}{\textbf{VidGuard-R1}}

\title{VidGuard-R1: AI-Generated Video Detection and Explanation via Reasoning MLLMs and RL}

\author{
Kyoungjun Park\textsuperscript{1}, 
Yifan Yang\textsuperscript{2}\thanks{Corresponding authors.}, 
Juheon Yi\textsuperscript{2}, 
Muhammad Muaz\textsuperscript{1}, 
Shicheng Zheng\textsuperscript{2}, \\
\textbf{Yifei Shen\textsuperscript{2}, 
Dongqi Han\textsuperscript{2}, 
Caihua Shan\textsuperscript{2}, 
Lili Qiu\textsuperscript{1,2}\footnotemark[1]}
\\[0.3em]
\textsuperscript{1}The University of Texas at Austin \quad \textsuperscript{2}Microsoft Research\\
\{kjpark, lili\}@cs.utexas.edu, yifanyang@microsoft.com
}

\iclrfinalcopy % Uncomment for camera-ready version, but NOT for submission.
\begin{document}

\maketitle
\vspace{-1.5em}

\begin{figure}[h]
% \vspace{-0.2in}
    \centering
    \includegraphics[width=1\linewidth]{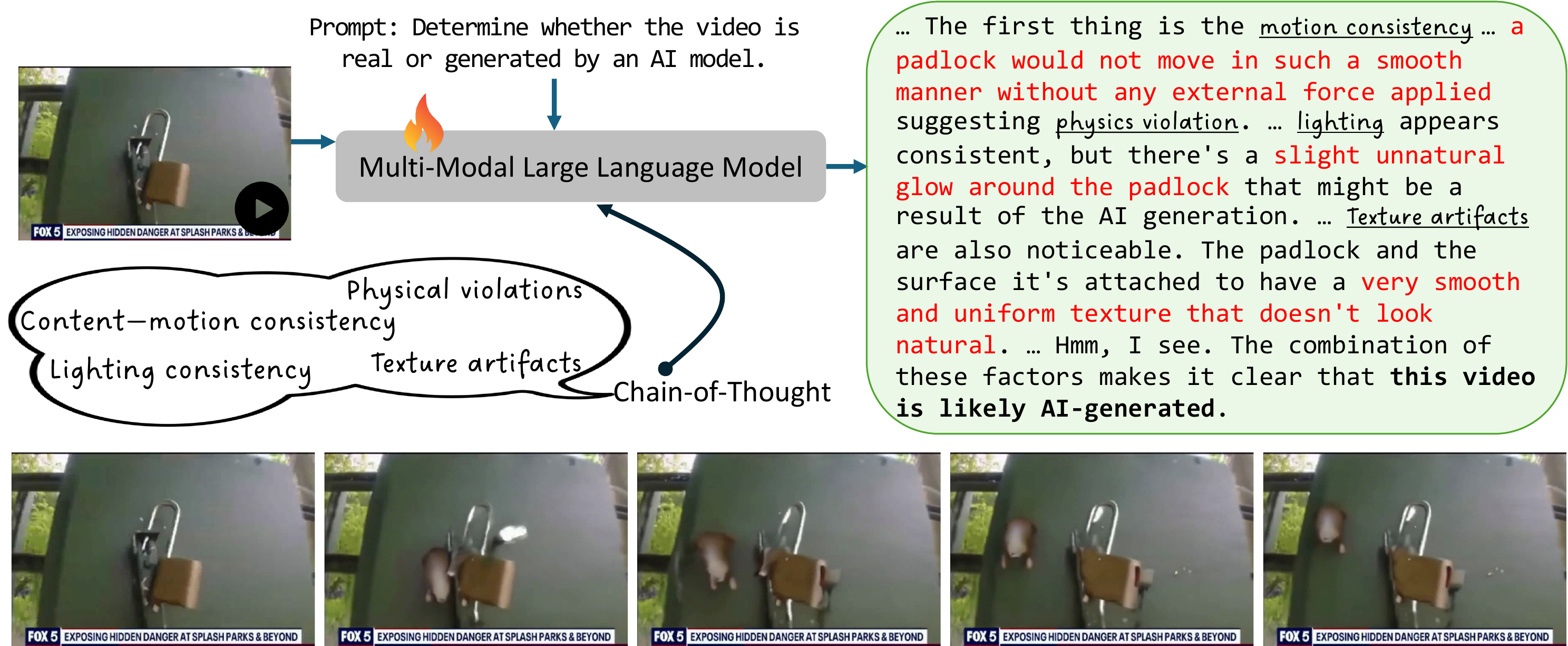}
    \vspace{-0.25in}
    \caption{Overall framework of \proj{}. We present the first video authenticity detector based on multi-modal large language models (MLLMs), which generates a chain-of-thought reasoning process along with the final answer.}
   \label{fig:teasure}
\end{figure}

\begin{abstract}
\vspace{-1em}
The rapid proliferation of AI-generated video necessitates robust detection tools that offer both high accuracy and human-interpretable explanations. While existing MLLM-based detectors rely on supervised fine-tuning (SFT) or direct preference optimization (DPO), these methods are often bottlenecked by static, pre-labeled datasets that fail to capture the evolving, multi-step physical inconsistencies of modern generative models. To bridge this gap, we introduce \proj{}, the first video authenticity detector to utilize group relative policy optimization (GRPO). Moving beyond passive preference matching, \proj{} employs a reinforcement learning framework that encourages the model to explore and rank multiple reasoning paths. By introducing specialized reward models for temporal stability and diffusion-aware complexity, we incentivize the model to discover 'physics-grounded' artifacts. Our contributions include: (1) a curated dataset of 140,000 challenging real/fake video pairs; (2) a GRPO-based training paradigm that achieves state-of-the-art zero-shot performance; and (3) a reasoning-first architecture that provides precise, verifiable rationales for its forensic judgments. Project website: \url{https://vidguard-r1.github.io/}

\end{abstract}

\vspace{-1em}
\section{Introduction}
\vspace{-0.5em}

In the past year, we have witnessed unprecedented progress in video generation models, with dramatic improvements in realism and quality. The release of powerful models such as Sora~\citep{videoworldsimulators2024}, Wan~\citep{wan2025}, and HunyuanVideo~\citep{kong2024hunyuanvideo} has made AI-generated videos more accessible to the public, further blurring the line between synthetic videos and real ones. At the same time, these advancements have led to a series of social risks, including the spread of misinformation, violations of privacy rights, damage to personal reputations, and increased susceptibility to scams and fraud.

%Therefore, there is urgent need to develop AI-generated video detectors that can accurately identify real and synthetic videos.

Motivated by its practical significance, several pioneering works have been developed to detect AI-generated videos. Early approaches primarily targeted DeepFake-style facial forgeries~\citep{qian2020thinking,npr,stil}, which often assumed single-subject, frontal-face scenarios under constrained settings. These assumptions diverge significantly from open-domain, multi-scene videos produced by modern generative models. More recent detectors leverage spatial-temporal consistency~\citep{decof,bai2024ai,liu2024turns}; however, such methods are limited in capturing higher-level semantic or causal inconsistencies and can be easily bypassed by post-processing techniques. Other methods are trained on curated fake video detection datasets~\citep{demamba,ni2025genvidbench,kundu2025towards}, but these benchmarks often lack coverage of newly emerging models and fail to reflect the full diversity of generative capabilities. A recent benchmark~\citep{demamba} shows that even state-of-the-art detectors still struggle to reliably identify videos from advanced models like Sora. Furthermore, these detectors typically offer only binary decisions without accompanying explanations, which raises concerns for transparency, especially when detection outcomes affect content moderation or legal accountability. Users are also more likely to trust detection systems that provide interpretable reasoning.

% Determining whether a video is AI-generated is a highly important application, as it can help prevent many societal risks. However, relying solely on another AI model to judge the authenticity of a video still carries the risk of hallucinations. Therefore, this paper argues that it is crucial to enable an AI video verification model to explain why a video is real or fake. Such explanations can assist users in making better informed judgments themselves. Similarly, inspired by works such as R1~\citep{video-r1}, this kind of reasoning and decision-making process can also lead to measurable improvements in the accuracy of video discrimination. 

Recent advances in multi-modal large language models (MLLMs) have significantly enhanced video understanding, enabling detailed explanations of model decisions~\citep{Qwen-VL,zhang2024video}. This makes them promising candidates for detecting and explaining AI-generated videos. However, directly applying existing MLLMs, including advanced models like GPT-4o, yields subpar performance on current benchmarks, underscoring the need for supervised fine-tuning (SFT). As an initial step, we applied SFT to the Qwen2.5-VL-7B model~\citep{Qwen2.5-VL}. While the model achieved strong overall performance, it remained limited in its ability to explain why a video is fake, revealing shortcomings in its reasoning capability.

To address this, we adopt reinforcement learning (RL), which has shown promise in enhancing LLM reasoning~\citep{guo2025deepseek}. Notably, Video-R1~\citep{feng2025video} outperforms commercial models on video reasoning tasks. RL enables MLLMs to develop self-improving reasoning via outcome-based rewards. We hypothesize that RL fine-tuning can help models detect subtle temporal and generative artifacts. Key to this is designing effective reward models. Simple binary rewards (e.g., 1 for real, 0 for fake) are insufficient. Instead, we propose two strategies: (1) injecting temporal artifacts into both real and fake videos to encourage temporal reasoning, and (2) assigning higher rewards to videos generated with more diffusion steps, which are harder to detect. Incorporating these into a group relative policy optimization (GRPO) framework yields over 86\% accuracy on our dataset and 95\% on two benchmarks.

\begin{itemize}
  
    \item We introduce \proj{}, the first video authenticity detector that fine-tunes the MLLM using GRPO. The model leverages the pretrained knowledge of MLLMs for accurate classification and employs reinforcement learning for effective exploration. To further enhance performance, we design two specialized reward models that target temporal artifacts and generation complexity based on diffusion steps.
    \item We construct a challenging dataset of 140k real/fake video pairs for AI-generated video detection. By employing state-of-the-art generation models and carefully controlling the process, we ensure that distinguishing real from fake is non-trivial.
    \item \proj{} achieves state-of-the-art zero-shot performance on existing benchmarks, with accuracy exceeding 95\%. Case studies further highlight its ability to produce accurate and interpretable explanations.

\end{itemize}

\vspace{-1em}
\section{Related works}\label{sec:related}  
\vspace{-0.5em}

\subsection{AI-generated video detection method}
\vspace{-0.5em}
Recent research on AI-generated video detection has largely focused on deepfake videos with synthetic faces~\citep{pei2024deepfake}, using spatial–temporal consistency, frequency artifacts, or data-driven approaches. These methods often struggle to generalize beyond face-centric content to more diverse, real-world videos. Recently, general video detection methods have emerged: AIGDet~\citep{bai2024ai} captures spatial–temporal anomalies, DeCoF~\citep{decof} exploits frame consistency, and diffusion-based representations track temporal dynamics~\citep{liu2024turns}. Other works identify appearance, motion, and geometry as key factors for classifier training~\citep{chang2024matters}.

Recent efforts in forgery detection leverage multimodal LLMs, including FakeShield~\citep{xu2024fakeshield}, which employs supervised fine-tuning, and SafeWatch~\citep{chen2024safewatch}, which applies direct preference optimization (DPO) for video guardrails. However, DPO relies on static preference pairs that struggle to capture subtle temporal inconsistencies in evolving generative models. We introduce the first application of group relative policy optimization (GRPO) to AI-generated video detection. By enabling iterative exploration and ranking of multiple reasoning paths, our method promotes a deeper physics-aware understanding of video consistency, leading to improved generalization across frontier generative models and diverse benchmarks.

% Multimodal LLMs have also been explored for forgery detection: FakeShield~\citep{xu2024fakeshield} uses supervised fine-tuning (SFT) for image forgery detection, while SafeWatch~\citep{chen2024safewatch} combines SFT and direct preference optimization (DPO) for video guardrails. In contrast, our work is the first to fine-tune a multi-modal LLM with group relative policy optimization (GRPO) for AI-generated video detection, demonstrating strong generalization across recent generative models and benchmark datasets.

\subsection{AI-generated video detection dataset}
\vspace{-0.5em}
Given the recency of this research area, only a limited number of benchmarks have been introduced. The generated video dataset (GVD)~\citep{bai2024ai} (11k samples) and GenVideo~\citep{demamba} (with millions of samples) consider settings where both training and test videos are generated by the same series of models. However, these benchmarks lack prompt/image–video pairs, semantic labels, or cross-source settings. GVF (2.8k samples) contains prompts/images–video pairs and semantic labels, but does not provide cross-source settings. GenVidBench~\citep{ni2025genvidbench} consists of 100k videos and incorporates cross-source settings, but the video generation models used are less advanced, such as CogVideo and SVD.

Moreover, existing datasets often contain shortcuts in resolution, frame rate, bitrate, or data-source imbalance, enabling models to exploit superficial cues rather than learn intrinsic visual realism. To address these limitations, we construct a curated dataset of 140{,}000 real–fake videos generated with state-of-the-art video generation models: HunyuanVideo~\citep{kong2024hunyuanvideo} and CogVideoX~\citep{yang2024cogvideox}. Our dataset explicitly standardizes bitrate, resolution, frame rate, and content distribution, resulting in a shortcut-free benchmark that encourages models to rely on semantic and temporal realism rather than on metadata artifacts.

\vspace{-0.5em}
\section{Methodology}
\vspace{-0.5em}

\begin{figure}[!t]
% \vspace{-0.2in}
    \centering
    \includegraphics[width=1\linewidth]{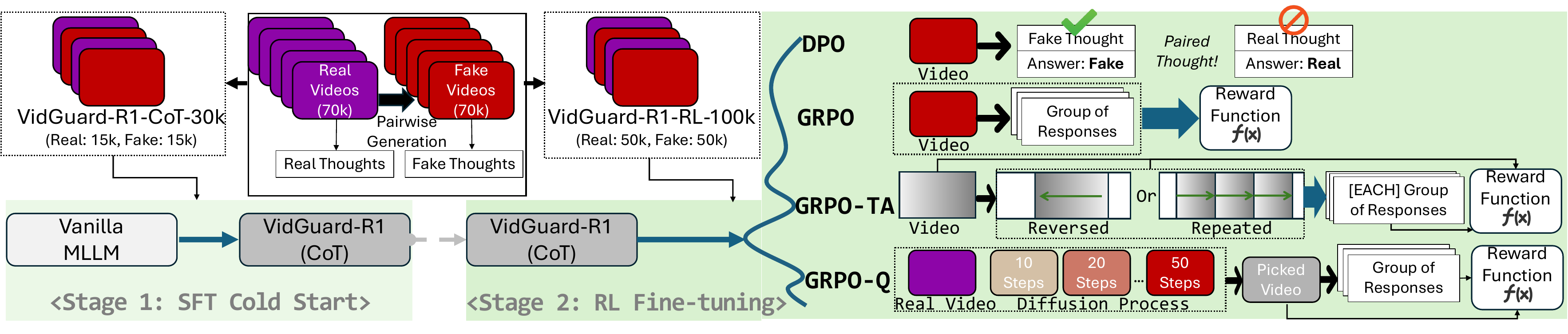}
    \vspace{-2em}
   \caption{The overall training framework of \proj{}, consisting of two stages: (1) supervised fine-tuning (SFT) for chain-of-thought (CoT) initialization, and (2) reinforcement learning-based fine-tuning to enable deeper reasoning.}
   \label{fig:overview}
   \vspace{-0.2in}
\end{figure}

Figure~\ref{fig:overview} illustrates the \proj{} framework, which consists of two stages. We first apply supervised fine-tuning (SFT) to the multimodal large language model (MLLM), followed by direct preference optimization (DPO) and group relative policy optimization (GRPO) based on the collected datasets. We further develop two GRPO variants by introducing temporal artifacts and leveraging videos generated with varying diffusion steps.

\vspace{-0.5em}
\subsection{Data collection} \label{sec:dataset_collection}
\vspace{-0.5em}

\subsubsection{Data construction for video realism discrimination}
\vspace{-0.5em}

% Activitynet 15k, InternVid 35k
% % why build our dataset? -> data curation part
High-quality training data is essential for video reasoning in MLLMs. However, many existing benchmarks for real vs. generated video classification, such as GenVideo~\citep{demamba} and GenVidBench~
\citep{ni2025genvidbench} exhibit uncontrolled discrepancies in basic metadata—e.g., real videos are often longer than 10 seconds, whereas generated videos are typically under 4 seconds in GenVideo. Moreover, they reveal clear modality-level gaps in motion dynamics and content contrasts between real and generated videos. These differences introduce unintended shortcuts, enabling models to rely on superficial cues like duration or resolution rather than actual visual realism. As a result, \proj{} attains over 96\% accuracy on both GenVideo and GenVidBench by exploiting such artifacts. To mitigate this reward hacking behavior, we construct a curated dataset with standardized video properties, encouraging models to focus on intrinsic visual content.

We collect real videos from the InternVid~\citep{wang2023internvid} and ActivityNet~\citep{caba2015activitynet} datasets and generate their corresponding fake counterparts using HunyuanVideo~\citep{kong2024hunyuanvideo} and CogVideoX~\citep{yang2024cogvideox}. We specifically choose these two models because they support conditioning on both the first-frame image and a text description—an essential requirement for generating videos that are contextually aligned with their real counterparts. To achieve such alignment, we provide the generation models with the first frame of each real video along with a textual caption describing its content. For ActivityNet, which lacks native captions, we extract concise descriptions using Qwen2.5-VL 72B. This pairing strategy mitigates content-based biases and forces the model to reason over subtle visual details.

\vspace{-0.5em}
\subsubsection{Collecting chain of thought (CoT) annotation}
\vspace{-0.5em}

Eliciting deliberate, step-by-step reasoning in MLLMs requires high-quality CoT supervision. To this end, we leverage Qwen-2.5-VL (72B) to extract salient visual cues from each video and guide the model toward a deeper understanding. Specifically, we query the model with critical factors known to distinguish real from generated content—motion consistency, lighting consistency, texture artifacts, and physical plausibility violations. These targeted prompts encourage detailed reasoning grounded in visual evidence.

However, current MLLMs lack the capacity to reliably distinguish real from fake videos on their own. To compensate, we provide ground-truth labels during prompt construction and instruct the model to generate CoT rationales conditioned on the given label. While these rationales do not reflect genuine discrimination ability, they capture rich contextual cues—such as object interactions, background details, and lighting inconsistencies—that are highly informative. These CoT annotations serve as useful clues for subsequent reinforcement learning fine-tuning. For prompt templates used in CoT generation, please refer to our supplementary materials.

\vspace{-0.5em}
\subsection{Supervised and RL Fine-tuning} \label{sec:standard_fine-tune}
\vspace{-0.5em}

We begin with SFT, where the model is trained to mimic the ground-truth reasoning process. Given a video $x$ and its annotation $y$ from the collected dataset, the model parameters $\theta$ are optimized by minimizing the negative log-likelihood $\mathcal{L}_{\text{SFT}}(\theta) = -\sum_{t=1}^{T} \log p_\theta(y_t \mid y_{<t}, x)$. To align the model outputs with human preferences, we apply DPO, which updates the model based on pairwise preference data without explicit reward modeling. Given a preferred response $y_w$ and a less-preferred response $y_l$ for the same video $x$, the DPO loss encourages the model to prefer $y_w$ over $y_l$ compared to a reference model $p_{\text{ref}}$:
\begin{small}
\begin{align}
\mathcal{L}_{\text{DPO}}(\theta) = 
& -\mathbb{E}_{(x,y_w,y_l)\sim D} \Bigg[
\log \sigma \Big(
\beta \log \frac{p_\theta(y_w|x)}{p_{\text{ref}}(y_w|x)} \nonumber 
 - \beta \log \frac{p_\theta(y_l|x)}{p_{\text{ref}}(y_l|x)}
\Big)
\Bigg]
\end{align}
\end{small}%
\leavevmode\noindent
where $\sigma(\cdot)$ is the sigmoid function and $\beta$ controls the preference strength. This method allows fine-tuning using preference comparisons without requiring scalar rewards.

Finally, we adopt GRPO from DeepSeek R1~\citep{guo2025deepseek}, which generalizes RLHF to group-level comparisons. Given a query video $x$ and a group of generated outputs $\{o_i\}_{i=1}^G$, the model is trained to assign higher probabilities to outputs with higher rewards. The GRPO objective is:
\resizebox{\linewidth}{!}{%
\begin{minipage}{\linewidth}
\begin{align}
\mathcal{L}_{\text{GRPO}}(\theta) = 
& -\mathbb{E}_{(x, o_{1:G}) \sim D} \Bigg[ 
\frac{1}{G}\sum_{i=1}^{G}
\min\Bigg(
\frac{p_{\theta}(o_i|x)}{p_{\text{ref}}(o_i|x)} A_i, \nonumber  \text{clip}\left(
\frac{p_{\theta}(o_i|x)}{p_{\text{ref}}(o_i|x)}, 
1-\epsilon, 1+\epsilon
\right) A_i 
\Bigg) 
- \beta D_{\text{KL}}\left(p_{\theta}\,\|\,p_{\text{ref}}\right) 
\Bigg]
\end{align}
\end{minipage}%
}
where $\epsilon$ is a clipping threshold and $\beta$ regularizes the policy to stay close to the reference model. The advantage term $A_i$ normalizes the reward $r_i$ for output $o_i$ within the group, computed as 
$A_i = \tfrac{r_i - \mu_x}{\sigma_x}$, 
where $\mu_x$ and $\sigma_x$ are the mean and standard deviation of $\{r_i\}_{i=1}^G$. GRPO thus enables learning from relative ranking among multiple responses, capturing nuanced distinctions in quality across outputs.

\vspace{-0.5em}
\subsection{ \proj{}} \label{sec:our_GRPO}
\vspace{-0.5em}
\subsubsection{Overview}
\vspace{-0.5em}

Figure~\ref{fig:overview} illustrates the training pipeline of \proj{}. Following the data collection procedure, we construct two datasets of different scales: \texttt{VidGuard-R1-CoT-30k} and \texttt{VidGuard-R1-RL-100k}. We adopt Qwen2.5-VL-7B as the base MLLM and train it using our proposed fine-tuning framework.

The first stage is supervised fine-tuning initialization using the \texttt{VidGuard-R1-CoT-30k} dataset, which contains videos paired with chain-of-thought (CoT) annotations. This stage establishes foundational reasoning ability and equips the model with basic cross-modal alignment and visual understanding. The resulting model is referred to as \textbf{VidGuard-R1 (CoT)}.

In the second stage, we apply two reinforcement learning methods—DPO and GRPO—to further refine the model on a larger and more diverse dataset, \texttt{VidGuard-R1-RL-100k}. DPO aligns the model with high-quality preference signals via pairwise comparisons, requiring the construction of preference pairs. Specifically, since our dataset includes pairwise real and fake videos, each sample is annotated with CoT rationales for both perspectives. For DPO training, we construct preference pairs by swapping these CoTs. For a real video, the CoT supporting its authenticity with the answer "real" serves as the positive annotation, while the CoT from the paired fake video with the answer "fake" is used as the negative annotation. In contrast, GRPO encourages consistent performance across grouped outputs by leveraging structural regularization. As it does not rely on preference annotations, video labels are directly used as reward signals. The resulting models are denoted as \textbf{VidGuard-R1 (DPO)} and \textbf{VidGuard-R1 (GRPO)}.

We introduce two variants, GRPO-TA and GRPO-Q, to further enhance detection performance. These methods extend the original GRPO framework by adjusting reward values according to the difficulty of detecting fake videos. Detailed descriptions are provided in the following sections.

% Figure~\ref{fig:overview} presents an overview of the training pipeline for \proj{}. 
% We adopt Qwen2.5-VL-7B as our base MLLM and train it in two stages: a supervised fine-tuning (SFT) initialization, followed by reinforcement learning (RL) optimization. This two-phase approach draws inspiration from recent advancements such as DeepSeek R1.

% In the first phase, we initialize the model with supervised fine-tuning on the \textsc{VidGuard-R1-CoT-20k} dataset, which consists of Chain-of-Thought (CoT) annotated samples generated by Qwen2.5-VL-72B. This stage establishes foundational reasoning capabilities and equips the model with basic cross-modal alignment and visual understanding. The model obtained from this step is referred to as the \textbf{SFT-CoT MLLM}.

% The second stage involves reinforcement learning on a larger and more diverse dataset, \textsc{VidGuard-R1-RL-70k}. We apply Direct Preference Optimization (DPO) and Group-wise Regularized Policy Optimization (GRPO) independently during this stage, training separate models with each method. While DPO focuses on aligning the model with high-quality preference signals through pairwise comparisons, GRPO emphasizes consistent performance across grouped outputs by leveraging structural regularization. The resulting models are denoted as \textbf{VidGuard-R1 (DPO)} and \textbf{VidGuard-R1 (GRPO)}, respectively.

\vspace{-0.5em}
\subsubsection{GRPO with temporal artifacts (GRPO-TA)}
\vspace{-0.5em}
While standard GRPO performs well in video discrimination by leveraging local visual cues—such as pixel distortions and lighting inconsistencies—it often overlooks temporal inconsistencies, which are crucial for detecting generated videos. To address this limitation, we introduce \textbf{GRPO with temporal artifacts (GRPO-TA)}, a variant that explicitly promotes temporal reasoning through a contrastive reward adjustment.

We apply two common temporal artifacts: (1) repeating a specific video segment and (2) reversing the frame sequence within a segment. These manipulations are applied probabilistically, with the manipulated region selected based on a Gaussian distribution over the video timeline.

Specifically, for each input query, we generate two sets of model outputs: $\{o_i\}_{i=1}^G$ for the original video, and $\{\tilde{o}_i\}_{i=1}^{G'}$ for the corresponding manipulated video with temporal artifacts. These videos should be classified as fake videos. In GRPO-TA, we assign additional rewards when the model correctly classifies temporally manipulated videos as fake. Consider two numbers, \(\alpha_1 > \alpha_2\). Detecting temporal artifacts in videos manipulated from real content tends to be more challenging than identifying those derived from fake videos. This is because real videos typically exhibit coherent and natural motion, so temporal manipulations such as frame shuffling or repetition can be subtle and difficult to detect. In contrast, generated videos often contain artifacts like unstable motion or low temporal consistency, which make further manipulations more visually salient. To reflect this asymmetry in difficulty, we assign the model a higher reward \(\alpha_1\) when the original video \(o_i\) is real, and a moderate reward \(\alpha_2\) when the original video is fake. This is defined as:
\begin{equation}
w_{i} =
\begin{cases}
\alpha_1, & \text{if } \tilde{o}_i = \textit{fake}  \text{ and } y_i = \textit{real} \\
\alpha_2, & \text{if } \tilde{o}_i = \textit{fake} \text{ and } y_i = \textit{fake} \\
0, & \text{otherwise}
\end{cases}
\end{equation}
\noindent
where \( y_i \) denotes the label of the $i$-th video. In the experiments, we set the hyperparameters to \( \alpha_1 = 0.5 \) and \( \alpha_2 = 0.3 \). This additional reward, $w_i$, is designed to be applied conditionally. Specifically, for a given sample, we only add $w_i$ to the original GRPO reward if two conditions are met: the model’s prediction on the original video ($O_i$) must be correct, and the overall accuracy on the group of manipulated videos ($\tilde{p}$) must exceed a predefined threshold $\mu$. This ensures that we only reward the model for temporal reasoning when it already has a solid baseline performance. The final reward of GRPO-TA is given by
\begin{equation}
r_i^{\text{GRPO-TA}} =
\begin{cases}
r_i^{\text{GRPO}} + w_i, & \text{if } o_i \text{ is correct and } \tilde{p} > \mu \\
r_i^{\text{GRPO}}, & \text{otherwise}
\end{cases}
\end{equation}
where \( r_i \) denotes the original GRPO reward, set to 1 if the model prediction on the original video is correct, and 0 otherwise. The additional reward \( w_i \) is applied only when both the original prediction is correct and the group of responses for the temporally manipulated videos achieves higher accuracy. In the experiments, we set $\mu = 0.8$.

% the ground-truth label (\textit{real} or \textit{fake}) of the \( i \)-th video.   Let \( \tilde{p} \) denote the accuracy on the manipulated set, measured as the proportion of correct predictions. We then incorporate $w_{i}$ into GRPO-TA: 

% \begin{equation}
% r_i^{\text{GRPO-TA}} =
% \begin{cases}
% r_i + w_i, & \text{if } o_i \text{ is correct and } \tilde{p} > \mu \\
% r_i, & \text{otherwise}
% \end{cases}
% \end{equation}
% \noindent
% \noindent
% where \( r_i \) denotes the original GRPO reward, set to 1 if the model prediction on the original video is correct, and 0 otherwise. The additional reward \( w_i \) is applied only when both the original prediction is correct and the group of responses for the temporally manipulated videos achieves higher accuracy. The parameter \( \mu = 0.8 \) controls the sensitivity to temporal artifacts, encouraging the model to prioritize temporal consistency and enhancing its robustness to subtle manipulations.

\vspace{-0.5em}
\subsubsection{GRPO with quality evolutionary videos (GRPO-Q)}
\vspace{-0.5em}
Our motivation is to extend the model’s capability to detect videos based on quality. Given the subjective nature of quality assessment, we avoid relying on large-scale human annotations. Instead, we leverage diffusion-based video generation by systematically varying the number of reverse diffusion steps to produce videos with distinct quality levels.

As in GRPO-TA, $o_i \in \mathcal{Y}$ and $y_i \in \mathcal{Y}$ denote the model output and ground-truth label, with 
$\mathcal{Y} = \{\text{real}\} \cup \{\text{fake-}s\}$, where $s$ is the diffusion step. A reward is given for an exact match, and no reward is assigned if the real/fake classification is incorrect. In GRPO-Q, if the model correctly classifies a fake video but selects an incorrect diffusion step, we assign a partial reward based on the distance between the predicted and ground-truth diffusion steps. The GRPO-Q reward is defined as follows:
\begin{equation}
r_i^{\text{GRPO-Q}} =
\begin{cases}
0, & \text{if } \big( o_i = \textit{real} \text{ and } y_i \neq \textit{real} \big) 
   \text{or } \big( o_i \neq \textit{real} \text{ and } y_i = \textit{real} \big)  \\
\delta, & \text{if } o_i = y_i \\
\left|g\left( o_i , y_i \right)\right|, & \text{if } o_i, y_i \in \mathcal{Y} \setminus \{ \textit{real} \}.
\end{cases}
\end{equation}
The first scenario occurs when the model fails to correctly classify the video as real or fake. The second scenario, where \(\delta = 1\), represents an exact match in prediction, including the diffusion progression. In the third case, the function \(g(\cdot, \cdot)\) maps the step distance to a scalar reward, enabling fine-grained credit assignment based on the similarity in quality. Specifically, we define a progress value \(s()\) in the range \([0, 1]\) to indicate the fraction of diffusion steps used, where \(0\) denotes zero steps, and \(1\) denotes full completion of the steps. The ground-truth value is \(s(y_i)\), and the model will estimate a progress value. We define the reward function as \(g(o_i, y_i) = \delta \cdot (1 - |s(o_i) - s(y_i)|)\).

% The model will output a guess of the ground-truth value $s(o_i)$ 

% In detail, $g()$ first extracts the diffusion progress values from $o_i$ and $y_i$, denoted as $s(o_i)$ and $s(y_i)$, respectively. It then computes a distance metric—such as the absolute difference $|s(o_i) - s(y_i)|$, and re-scale and reverse the distance into the range $[0, \delta]$. The reward is higher when the predicted and ground-truth diffusion steps are similar, and it decreases linearly as their distance grows.
%The output of \( g(\cdot) \) is constrained to be less than \( \delta \), ensuring that partial rewards remain lower than fully correct ones.

This reward formulation enables the model to move beyond binary discrimination and perform fine-grained analysis of video quality. By learning to associate subtle differences in generation steps with quality variations, the model develops a deeper understanding of the diffusion process and its impact on perceptual realism. As a result, it can not only detect whether a video is fake, but also infer and estimate the degree of quality degradation in generated videos. This facilitates more interpretable and controllable evaluation of generated content quality.

\vspace{-0.5em}
\section{Experiments}
\vspace{-0.5em}

\subsection{Implementation details}
\vspace{-0.5em}

\subsubsection{Dataset}
\vspace{-0.5em}
Our dataset contains 140k videos, balanced between 70k real and 70k generated samples, organized into contextual pairs. The real set comprises 55k videos from InternVid and 15k from ActivityNet, while the generated set includes 50k samples synthesized by HunyuanVideo-I2V~\citep{kong2024hunyuanvideo} and 20k by CogVideoX-5B~\citep{yang2024cogvideox}. We allocate 130k samples for training and 10k for testing, with the latter evenly split between real and generated videos. Within the training data, 30k samples are reserved for chain-of-thought (CoT) learning, denoted as \texttt{VidGuard-R1-CoT-30k}, and the remaining 100k are used for reinforcement learning fine-tuning, denoted as \texttt{VidGuard-R1-RL-100k}.  

% first application of discriminating fake/real: fake points, real points using GRPO
% 4 metrics,

Since the state-of-the-art generative models still produce relatively short videos ($\sim$129 frames) at modest resolutions, we standardize all real videos to match generated ones by enforcing 49 frames, 8 FPS, 720$\times$480 resolution, and YUV420p format.  

For GRPO-Q fine-tuning, we augment the training set with intermediate generations sampled from diffusion steps 10 to 50. These are labeled with approximate quality levels (20\%, 40\%, 60\%, 80\%, and 95\%). Specifically, we use 12k real videos, each paired with five generated variants at different diffusion steps, resulting in 72k samples per generation model.

\vspace{-0.5em}
\subsubsection{Evaluation protocol}
\vspace{-0.5em}
We evaluate three datasets—ours, GenVidBench~\citep{ni2025genvidbench}, and GenVideo~\citep{demamba}—using the metrics and baselines defined by their respective benchmarks. For ours and GenVidBench, we report \textbf{mean Top-1 accuracy}, the average correctness over all predictions. For GenVideo, we follow the original protocol and report \textbf{recall} and \textbf{F1 score}. All evaluations adhere to the official settings of each benchmark to ensure fair comparison.

\vspace{-0.5em}
\subsubsection{Training Setup}
\vspace{-0.5em}
We employ Qwen2.5-VL-7B as the base MLLM and conduct all experiments on four NVIDIA A100 GPUs (80GB). Each video is represented by up to 16 frames, each resized to $28 \times 28$ and mapped to 128 feature channels for encoder input during both training and inference. For GenVideo and GenVidBench, we follow their official evaluation protocols and adopt 8-frame inputs. In GRPO training, we sample 8 responses per input; for GRPO-TA, we additionally sample 4 responses from temporally manipulated variants of the input to enhance robustness against temporal artifacts. Training proceeds in two stages: first, the base model is fine-tuned for one epoch on the CoT dataset, yielding the SFT-CoT MLLM; second, we initialize \proj{} with SFT-CoT and perform reinforcement learning for approximately 2,000 steps.

% This mechanism encourages informative yet concise reasoning by rewarding outputs whose lengths fall within a target range. Specifically, if the model predicts the correct answer and the response length falls within \([l_{\min}, l_{\max}]\), an additional reward \( \omega \) is assigned:

% We follow the original configuration with \( \omega = 0.1 \), \( l_{\min} = 320 \), and \( l_{\max} = 512 \), encouraging the model to produce rationales that are neither too brief nor excessively long.

% how many frames
% 128 × 28 × 28 -> 256 x 28 x 28 during inf
% SFT 1 epoch, GRPO takes 2000 steps
% 16 frames for my dataset
% 8 frames for genvideo and genvidbench bs they report their benchmark

\begin{table}[t]
\centering
\resizebox{\linewidth}{!}{%
\begin{minipage}{0.4\textwidth} % 왼쪽 테이블
\centering
\caption{Comparison of models on our dataset, reported as mean Top-1 accuracy (\%). TF denotes transformer.}
    \label{tab:our}
\resizebox{1\textwidth}{!}{%
\begin{tabular}{ccccc}
\toprule
\textbf{Method} & \textbf{Type} & \textbf{CogVideoX} & \textbf{HunyuanVideo} \\
\hline
SlowFast & CNN & 77.87  & 77.03 \\
I3D & CNN  & 64.78  & 62.13 \\
TRN & CNN  & 68.73 & 69.87 \\
\hline 
UniFormer V2 & TF & 73.95  & 71.92\\
TimeSformer & TF  & 78.53  & 74.55 \\
VideoSwin & TF  & 76.81   & 79.71 \\
MViT V2 & TF  & 58.38  & 53.91 \\
\hline
\midrule
Qwen2.5-VL-7B & MLLM & 50.95 & 52.83 \\
Qwen2.5-VL-72B & MLLM & 54.17 & 55.82 \\

GPT-4.1 mini & MLLM & 54.95 & 55.31 \\
GPT-4o & MLLM & 56.81 & 57.42 \\
VidGuard-R1 (CoT) & MLLM & 66.18 & 63.19 \\
VidGuard-R1 (DPO) & MLLM & 79.13  & 80.88 \\
VidGuard-R1 (GRPO) & MLLM & 81.30  & 81.90 \\
VidGuard-R1 (GRPO-TA) & MLLM & 82.17  & 83.72 \\
VidGuard-R1 (GRPO-Q) & MLLM &  \textbf{84.32}   & \textbf{86.17}\\
\hline
\end{tabular}%
}
\end{minipage}%

\hspace{0.03\textwidth}%

\begin{minipage}{0.55\textwidth} % 오른쪽 테이블
\centering
\caption{Extended GenVidBench results with \proj{} and additional MLLMs, reported as mean Top-1 accuracy (\%). TF denotes transformer.}
    \label{tab:genvidbench}
\resizebox{1\textwidth}{!}{%
\begin{tabular}{ccccccc|c}
\toprule
\textbf{Method} & \textbf{Type} & \textbf{MuseV} & \textbf{SVD} & \textbf{CogVideo} & \textbf{Mora} & \textbf{HD-VG} & \textbf{Mean} \\
\hline
SlowFast & CNN& 12.25  & 12.68  & 38.34  & 45.93  & 93.63  & 41.66  \\
I3D& CNN& 8.15  & 8.29  & 60.11  & 59.24  & 93.99  & 49.23  \\
TRN & CNN& 38.92  & 26.64  & 91.34  & 93.98  & 93.97  & 71.26  \\
\hline 
UniFormer V2 & TF & 20.05  & 14.81  & 45.21  & 99.21 & 96.89  & 57.55  \\
TimeSformer & TF & 73.14  & 20.17  & 74.80  & 39.40  & 92.32  & 64.28  \\
VideoSwin & TF & 62.29  & 8.01  & 91.82 & 45.83  & \textbf{99.29} & 67.27  \\
MViT V2 & TF & 76.34  & \textbf{98.29} & 47.50  & 96.62  & 97.58  & 79.90 \\
\hline 
\midrule

Qwen2.5-VL-7B & MLLM & 25.86  &  27.06  & 68.51  & 43.26  & 71.15 & 47.30 \\
GPT-4.1 mini & MLLM & 26.07  & 33.78 & 94.07  & 57.19  & 87.64  & 59.62 \\
VidGuard-R1 (CoT) & MLLM& 36.52  & 16.02 & 99.35  & 76.94  & 99.94  &66.09 \\
\makecell[c]{VidGuard-R1 (GRPO,\\ GenVideo-pretrained,\\Zero-shot)}
& MLLM&  97.24 & 96.59 & 99.88  & 99.93  & 88.14  & 96.37 \\
VidGuard-R1 (GRPO) & MLLM & \textbf{97.38}  & 94.98 & \textbf{99.90}  & \textbf{99.99}  & 95.46  & \textbf{97.53} \\
\hline
\end{tabular}%
}
\end{minipage}
}
\vspace{-1em}
\end{table}

\vspace{-0.5em}
\subsection{Main results}
\vspace{-0.5em}
\subsubsection{Our dataset}
\vspace{-0.5em}
% much harder than two benchmarks / ex  -> better understanding fake feature withoutxx
\iffalse

\begin{table}[!t]

  \centering
  \resizebox{0.5\textwidth}{!}{
  \begin{tabular}{ccccc}
 \toprule
 \textbf{Method} & \textbf{Type} & \textbf{Mean Top-1} & \textbf{F1} \\
 \hline
 SlowFast~\citep{slowfast} & CNN & 77.87  & 0.81\\
 % F3Net~\citep{f3net} & CNN & x  & x  & x\\
 I3D~\citep{i3d} & CNN  & 64.78  & 0.66 \\
 % CFV2~\citep{cfv2} & CNN & x  & x  & x\\
 % TPN~\citep{tpn} & CNN & x  & x  & x\\
 % TIN~\citep{tin} & CNN & x  & x  & x\\
 TRN~\citep{trn} & CNN  & 68.73 & 0.69\\
 % TSM~\citep{tsm} & CNN & x  & x  & x\\
 % X3D~\citep{x3d} & CNN  & x  & x  & x\\
 \hline 
 UniFormer V2~\citep{uniformerv2} & Transformer & 73.95  & 0.76\\
 TimeSformer~\citep{timesformer} & Transformer  & 78.53  & 0.80 \\
 VideoSwin~\citep{videoswin} & Transformer  & 76.81   & 0.80 \\
 MViT V2~\citep{mvit} & Transformer  & 58.38  & 0.57 \\
 \hline 
 \midrule
 Qwen2.5-VL-7B~\citep{Qwen2.5-VL} & MLLM & 50.95 & 0.67\\
 GPT-4.1 mini~\citep{gpt4} & MLLM & 54.95 & 0.35 \\
 VidGuard-R1 (CoT) & MLLM & 66.18 & 0.79\\
 VidGuard-R1 (DPO) & MLLM & 79.13  & 0.80\\
 VidGuard-R1 (GRPO) & MLLM & 81.30  & 0.81\\
 VidGuard-R1 (GRPO-TA) & MLLM & 82.17  & 0.83\\
 VidGuard-R1 (GRPO-Q) & MLLM &  \textbf{84.32}   & \textbf{0.86}\\
 \hline
  \end{tabular}
  
  }
 \caption{Performance comparison of various models on our dataset}
\label{tab:our}
\end{table}
\fi

We evaluate \proj{} on our dataset with several methods, including CNN-based models (SlowFast~\citep{slowfast}, I3D~\citep{i3d}, TRN~\citep{trn}), Transformer-based models (UniFormer V2~\cite {uniformerv2}, TimeSformer~\citep{timesformer}, VideoSwin~\citep{videoswin}, MViT V2~\citep{mvit}), and MLLM-based models (Qwen2.5-VL~\citep{Qwen2.5-VL} and GPT-4.1 mini~\citep{gpt4}). For CNN and Transformer models, we use the default training settings provided by the MMAction2 framework~\citep{2020mmaction2}.

As shown in Table~\ref{tab:our}, CNN- and Transformer-based models achieved 53–79\% accuracy, with SlowFast and TimeSformer among the top performers. In contrast, Qwen2.5-VL-7B and GPT-4.1 mini exhibited near-random performance, highlighting their limited capability in distinguishing fake videos. \textit{VidGuard-R1 (CoT)}, trained via supervised fine-tuning (SFT) on Qwen2.5-VL-7B, substantially improved accuracy from around 51\% to over 66\%, yet remained less competitive compared to advanced SOTA methods. This result aligns with the intended role of the SFT stage---as a cold start phase to guide the model toward structured \textit{think + answer} responses, emphasizing the extraction of salient visual cues.

In the subsequent RL stage, both DPO and GRPO further improved performance by roughly 2\% over the best baseline. Our proposed methods---GRPO-TA and GRPO-Q---achieved additional gains of approximately 2\% and 5\% over GRPO, respectively, demonstrating the effectiveness of temporal artifact supervision and quality-aware reward modeling in enhancing detection accuracy.

\vspace{-0.5em}
\subsubsection{GenVidBench benchmark}
\vspace{-0.5em}
The GenVidBench dataset comprises approximately 87k training samples and 82k testing samples, with fake videos generated by models such as MuseV~\citep{musev}, SVD~\citep{svd}, CogVideo~\citep{cogvideo}, and Mora~\citep{mora}, and real videos sourced from HD-VG~\citep{hd_vg_130m}. We conduct training and evaluation under the cross-source and cross-generator settings as proposed in their benchmark. In addition to the models originally reported in GenVidBench, we evaluate \proj{} using the same model families as in our dataset experiments—CNN-based, Transformer-based, and MLLM-based models—including two MLLMs: Qwen2.5-VL and GPT-4.1 mini. \texttt{VidGuard-R1 (GRPO, GenVideo-pretrained, Zero-shot)} denotes the zero-shot model pretrained on GenVideo and evaluated on GenVidBench. As shown in Table~\ref{tab:genvidbench}, both the zero-shot model and two fine-tuned variants achieve over 15\% higher accuracy compared to prior SOTA methods. Notably, the zero-shot model demonstrates strong generalization, highlighting the effectiveness of pretraining on diverse generative content. Complete detection model results are provided in Appendix~\ref{sec:add_eval}.

\begin{table*}[t]
    \centering
    \renewcommand{\arraystretch}{0.8} % 행간 통일

    \caption{Extended GenVideo results with \proj{} and additional MLLMs, evaluated by F1 score and recall (R)}
    
    \label{tab:genvideo}
    \resizebox{\textwidth}{!}{
    \begin{tabular}{ccccccccccccc|c}
    \toprule
       \multirow{2}{*}{\textbf{Method}} & \textbf{Detection}&\multirow{2}{*}{\textbf{Metric}}&\multirow{2}{*}{\textbf{Sora}}&Morph&\multirow{2}{*}{\textbf{Gen2}}&\multirow{2}{*}{\textbf{HotShot}}&\multirow{2}{*}{\textbf{Lavie}}&\multirow{2}{*}{\textbf{Show-1}}&\textbf{Moon}&\multirow{2}{*}{\textbf{Crafter}}&\textbf{Model}&\textbf{Wild}&\multirow{2}{*}{\textbf{Mean}}\\
       &\textbf{level}&&&\textbf{Studio}&&&&&\textbf{Valley}&&\textbf{Scope}&\textbf{Scrape}\\

        \midrule
       \multirow{2}{*}{NPR~\citep{npr}}&\multirow{2}{*}{Image}&
R& 0.91 &0.99& 0.99& 0.24& 0.89& 0.57 &0.97
       &0.99&0.94&0.87&0.84\\
       &&F1&0.27&0.84&0.91&0.30&0.86&0.59&
0.81&0.91&0.81&0.81&0.71
       \\
       
       \midrule

       \multirow{2}{*}{VideoMAE~\citep{videomae}}&\multirow{2}{*}{Video}&
        R&
       0.67 & 0.96 &0.98 & 0.96& 0.77 & 0.80 &0.97&0.96& 0.96
       &0.68&0.87\\
        &&F1&0.62&0.95&0.98&\textbf{0.96}&0.86&0.87&0.96&0.97&\textbf{0.96}&0.79&0.89
       \\

       \midrule       
\multirow{2}{*}{MINTIME-CLIP~\citep{minetime}}&\multirow{2}{*}{Video}&
R& 0.89&1.00 &0.98 &0.26 &0.96&0.98 &0.99 &  1.00&0.84 &0.82 &0.87
\\
&&F1&0.49 &0.93 & 0.96 &0.37&0.94 &0.92 &0.92 & 0.96& 0.84&0.85 &0.82\\

\midrule
       \multirow{2}{*}{FTCN-CLIP~\citep{ftcn}}&\multirow{2}{*}{Video}&
R&0.87 & 1.00 & 0.98 & 0.17 & 0.97 & 0.91 & 1.00 & 1.00 &  0.85& 0.82& 0.86 \\
&&F1& 0.78 & 0.98 & 0.98 & 0.29 & 0.98 & 0.94 & 0.98 & \textbf{0.99} & 0.90 & 0.89 & 0.87\\

\midrule

       \multirow{2}{*}{DeMamba-XCLIP~\citep{demamba}}&\multirow{2}{*}{Video}&
        R
       &0.98& 1.00 & 0.99 & 0.65 & 0.94 & 0.98 & 1.00& 1.00&0.92 &0.89 & 0.93\\
       &&F1&0.64& 0.96& 0.97& 0.75& \textbf{0.95}&0.95&0.95&0.97&0.92&\textbf{0.91}&0.90
        \\
        \midrule
       \midrule
        \multirow{2}{*}{Qwen2.5-VL-7B~\citep{Qwen2.5-VL}}&\multirow{2}{*}{MLLM}&
        R&
       0.58 &0.56 &0.54 & 0.33 & 0.43 &0.38 &0.81 &0.63 & 0.51 & 0.70 &0.54 \\
       &&F1&0.74 &0.72 &0.70 &0.49 &0.60 &0.55 &0.90 &0.77 &0.68 &0.82 &0.70 \\
       \midrule
        \multirow{2}{*}{GPT-4.1 mini~\citep{gpt4}}&\multirow{2}{*}{MLLM}&
        R&
       0.43 &0.67  &0.56  & 0.54  & 0.63 &0.56  &0.92 &0.67 & 0.69  & 0.69 &0.65 \\
       &&F1&0.60 &0.80 &0.72&0.70&0.77 &0.72 &0.96 &0.80 &0.82 &0.82 &0.72 \\
       \midrule
        \multirow{2}{*}{VidGuard-R1 (CoT)}&\multirow{2}{*}{MLLM}& 
        R&
       0.92 & 0.89 & 0.91 & 0.90 & 0.98 & 0.79 &0.99 &0.85  & 0.89  & 0.87 &0.90 \\
       &&F1&0.90 &0.91 &0.95 &0.89 &0.99 &0.81 & 0.95 & 0.89& 0.85&0.88 & 0.90 \\
       \midrule
       \multirow{2}{*}{\makecell[c]{VidGuard-R1 (GRPO,\\ GenVidBench-pretrained, Zero-shot)}} & \multirow{2}{*}{MLLM} &
        R&
       0.95 &0.98 &0.90 & 0.89 & 0.97 & 0.85 & 0.99 & 0.93 & 0.81 & 0.87 & 0.92\\
       && F1 & 0.93& 0.93 & 0.96 & 0.91 & 0.99 & 0.82 & 0.95 & 0.89& 0.85& 0.88 & 0.91\\

       \midrule
        \multirow{2}{*}{VidGuard-R1 (GRPO)}&\multirow{2}{*}{MLLM}&
        R&
       0.95&1.00 &0.98 & 0.94& 0.98 & 0.95 & 0.97 & 0.99 & 0.94 & 0.91 & 0.96\\
       && F1 & \textbf{0.97} & \textbf{0.99} & \textbf{0.99} &0.91 & \textbf{0.99} & 0.89 & \textbf{0.99} & \textbf{0.99} & 0.95& 0.90& \textbf{0.96}\\

        \bottomrule
    \end{tabular}}
\vspace{-1em}
\end{table*}

\vspace{-0.5em}
\subsubsection{GenVideo benchmark}
\vspace{-0.5em}

The GenVideo dataset comprises approximately 2.2M training samples and 20k testing samples, with generated videos sourced from a diverse set of models, including Sora~\citep{sora2025}, MorphStudio~\citep{morph}, Gen2~\citep{gen2}, HotShot~\citep{hotshot}, Lavie~\citep{lavie}, Show-1~\citep{show1}, MoonValley~\citep{moonvalley}, Crafter~\citep{crafter}, ModelScope~\citep{modelscope}, and WildScrape~\citep{dreamvideo}. Following the official evaluation protocol, we benchmark two MLLM baselines and three variants of \proj{}. Among these, \texttt{VidGuard-R1 (GRPO)} consistently outperforms almost all prior detection methods across videos generated by the various models. As shown in Table~\ref{tab:genvideo}, it achieves an F1 score improvement of 0.06 compared to DeMamba-XCLIP. Complete detection model results are provided in Appendix~\ref{sec:add_eval}.

\vspace{-0.5em}
\subsubsection{Performance gap between our dataset and benchmarks}
\label{subsubsec:gap}
\vspace{-0.5em}

While \proj{} achieves approximately 85\% accuracy on our dataset, it obtains significantly higher accuracy—exceeding 95\%—on the two benchmark datasets. This discrepancy arises from two key differences. First, the benchmarks exhibit clear discrepancies in video metadata—such as resolution, duration, and frame rate—between real and fake videos, which models can exploit as superficial cues. In contrast, we standardize all videos in our dataset by matching resolution, FPS, and format, thereby forcing models to rely on actual visual content. Second, our dataset ensures strong contextual alignment by conditioning generation on the first frame and the corresponding caption of a real video, resulting in more realistic and semantically consistent outputs. In comparison, benchmark datasets often generate fake videos from unrelated prompts and synthetic images, leading to artifacts that make detection easier.

\vspace{-0.5em}
\subsubsection{Ablation study}
\vspace{-0.5em}
% In the Appendix, we assess explanation quality using an LLM-as-a-judge with GPT-4.1 mini. We show that the explanation quality consistently improves through RL fine-tuning. We also include ablation studies on the hyperparameters of GRPO-TA and the number of diffusion steps used in GRPO-Q.

\begin{wraptable}{r}{0.38\textwidth}
\vspace{-1.5em}
\centering
\caption{LLM-as-a-judge explanation scores on our dataset}
  \resizebox{0.38\textwidth}{!}{

\begin{tabular}{lccc}
  \toprule
  \textbf{Method} & \makecell{\textbf{Expl. Score}\\\textbf{(HunyuanVideo)}} & \makecell{\textbf{Expl. Score}\\\textbf{(CogVideoX)}} \\
  \midrule
  Qwen2.5-VL-7B & 5.8 & 5.6 \\
  GPT-4.1 mini & 5.8 & 5.9 \\
  VidGuard-R1 (CoT) & 6.8 & 6.9 \\
  VidGuard-R1 (DPO) & 7.2 & 8.1 \\
  VidGuard-R1 (GRPO) &  8.1 & 8.0 \\
  VidGuard-R1 (GRPO-TA) &  8.1& 8.4 \\
  VidGuard-R1 (GRPO-Q) & \textbf{8.3} & \textbf{8.5} \\
  \bottomrule
\end{tabular}
}

\label{tab:llm-as-judge}
% \vspace{-0.7em}
\end{wraptable}

\paragraph{Explanation quality and accuracy comparison.}  
Table~\ref{tab:llm-as-judge} presents results on the HunyuanVideo~\citep{kong2024hunyuanvideo} and CogVideoX~\citep{yang2024cogvideox} datasets. We report explanation quality scores, which are rated on a 1–10 scale (with 10 indicating excellent quality and full alignment) by GPT-4.1 mini using the LLM-as-a-judge prompt described in Appendix~\ref{sec:prompt}. Compared to baseline models such as Qwen2.5-VL-7B and GPT-4.1 mini, our VidGuard-R1 GRPO variants achieve consistent improvements in both classification accuracy and explanation quality.

\begin{wraptable}{r}{0.35\textwidth}
\vspace{-3em}
\centering
\caption{Accuracy (\%) for \textbf{GRPO-TA} under different reward function parameters 
$\alpha_1$ and $\alpha_2$}
  \resizebox{0.17\textwidth}{!}{

\begin{tabular}{cc|c}
\toprule
$\alpha_1$ & $\alpha_2$ & Accuracy (\%) \\
\midrule
0.3 & 0.1 & 81.31 \\
0.3 & 0.3 & 82.59 \\
\textbf{0.5} & \textbf{0.3} & \textbf{83.57} \\
0.5 & 0.5 & 83.12 \\
0.7 & 0.5 & 82.53 \\
\bottomrule
\end{tabular}
}
\vspace{-2em}
\label{tab:ablation-alpha}
\end{wraptable}

\paragraph{GRPO-TA reward ablation.}
Table~\ref{tab:ablation-alpha} reports an ablation study of GRPO-TA on our dataset by varying the weight parameters $\alpha_1$ and $\alpha_2$, which control the relative importance of different temporal artifact types. The highest classification accuracy of 83.57\% is achieved with $\alpha_1=0.5$ and $\alpha_2=0.3$, while the threshold $\mu$ is fixed at 0.8 across all experiments.

\begin{table*}[t]
\centering
\begin{minipage}[t]{0.35\textwidth}
\centering
\caption{Accuracy (\%) for \textbf{GRPO-Q} with varying number of intermediate diffusion steps}
\vspace{0.4em} 
\label{tab:ablation-grpoq-steps}
\resizebox{\textwidth}{!}{
\begin{tabular}{c|c}
\toprule
\# of steps (step numbers) & Accuracy (\%) \\
\midrule
1 (50) & 81.63 \\
3 (10, 30, 50) & 83.21 \\
\textbf{5 (10, 20, 30, 40, 50)} & \textbf{85.80} \\
\bottomrule
\end{tabular}
}
\end{minipage}
\hfill
\begin{minipage}[t]{0.58\textwidth}
\centering

\caption{Cross-dataset evaluation for \proj{}}
\vspace{0.4em}
\label{tab:mixing_results}
\resizebox{\textwidth}{!}{
\begin{tabular}{l l c}
\toprule
\textbf{Test Dataset} & \textbf{Training Source} & \textbf{Accuracy (\%)} \\
\midrule
Ours       & \proj{} (Ours only)         & 81.65 \\
Ours       & \proj{} (Ours + GenVideo)   & 82.97 \\
GenVideo   & \proj{} (GenVideo only)     & 97.53 \\
GenVideo   & \proj{} (Ours + GenVideo)   & 97.98 \\
\bottomrule
\end{tabular}

}
\end{minipage}
\end{table*}

\paragraph{GRPO-Q reward ablation.}
Table~\ref{tab:ablation-grpoq-steps} presents an ablation study on GRPO-Q conducted on our dataset by varying the number of intermediate diffusion steps included per real video during fine-tuning. Using more steps provides richer supervision of video quality progression, improving detection accuracy. The best accuracy of 85.80\% is obtained with five steps, which is the setting used in our main experiments.

\paragraph{Cross-dataset complementarity.}
To assess whether training on a limited generative source induces overfitting, we conduct dataset-mixing experiments using \proj{} (GRPO). As shown in Table~\ref{tab:mixing_results}, augmenting our dataset with GenVideo leads to consistent performance gains across both evaluation sets, suggesting that the model benefits from heterogeneous training data and does not rely on artifacts from any single source. These findings indicate that incorporating diverse generative sources enhances overall accuracy, reinforcing that \proj{} learns generalizable detection cues rather than dataset-specific patterns.

\subsection{Human Evaluation of Explanation Quality}

To examine the coherence and interpretability of CoT rationales, we conducted a human evaluation of \proj{} (GRPO). Tables~\ref{tab:consistency-quality} and \ref{tab:ranking-study} summarize the two complementary studies.

% ---------------------------------------------------------
% TABLES SIDE BY SIDE
% ---------------------------------------------------------
\begin{table*}[t]
\centering
\small
\begin{minipage}[t]{0.48\linewidth}
    \centering
    \caption{Label--rationale consistency and explanation quality for 20 videos}
    \label{tab:consistency-quality}
    \vspace{0.4em}
    \begin{tabular}{l c}
        \toprule
        \textbf{Metric} & \textbf{Value} \\
        \midrule
        Annotators & 5 \\
        Label–rationale alignment & 89\% \\
        Rationale score (0--5) & 3.9 \\
        \bottomrule
    \end{tabular}
\end{minipage}
\hfill
\begin{minipage}[t]{0.48\linewidth}
    \centering
    \caption{Human ranking of explanation quality}
    \label{tab:ranking-study}
    \vspace{0.4em}
    \begin{tabular}{l c}
        \toprule
        \textbf{Model} & \textbf{Avg. Rank} \\
        \midrule
        \proj{} (GRPO) & 1.67 \\
        GPT-4o & 2.08 \\
        Qwen2.5-VL-72B & 2.22 \\
        \bottomrule
    \end{tabular}
\end{minipage}
\end{table*}
% ---------------------------------------------------------
% TEXT
% ---------------------------------------------------------

\paragraph{Consistency and quality.}
Five annotators evaluated twenty randomly selected fake videos that the model correctly identified. Annotators judged whether each rationale was consistent with the predicted label and assigned a quality score on a 0--5 scale after watching the corresponding video. As shown in Table~\ref{tab:consistency-quality}, annotators reported 89\% label--rationale agreement with an average quality score of 3.9. Lower scores ($\leq 2$) occurred primarily when the model emphasized subtle texture-level artifacts that were difficult for human raters to perceive.

\paragraph{Explanation quality ranking.}
We conducted a blind ranking study in which five participants evaluated explanations from three models across twenty videos, consisting of ten real and ten synthetic videos. Table~\ref{tab:ranking-study} shows that \proj{} achieved the strongest human preference, with the lowest average rank (1.67), outperforming GPT-4o and Qwen2.5-VL-72B. Participants consistently judged \proj{}’s explanations to be more informative and higher quality. This improvement is likely because, although GPT-4o and Qwen2.5-VL-72B also describe video features well, their explanations are less effective at distinguishing real from synthetic content, whereas \proj{} better highlights the discriminative aspects relevant to this task.

\paragraph{Interpretation and broader implications.}
Although the CoT rationales align closely with human judgments, they are not guaranteed to be correct in every case. Hallucinations may arise, and the rationales should therefore be shown with an appropriate disclaimer noting their AI-generated nature. Nevertheless, presenting interpretable cues helps users more effectively assess whether a video may be AI-generated, supporting a practical human-in-the-loop verification process.

\vspace{-0.5em}

\subsection{Case studies on explanations}
\vspace{-0.5em}
\begin{figure*}[!t]
\centering
\begin{minipage}[t]{0.49\textwidth}
    \centering
    \includegraphics[width=\linewidth]{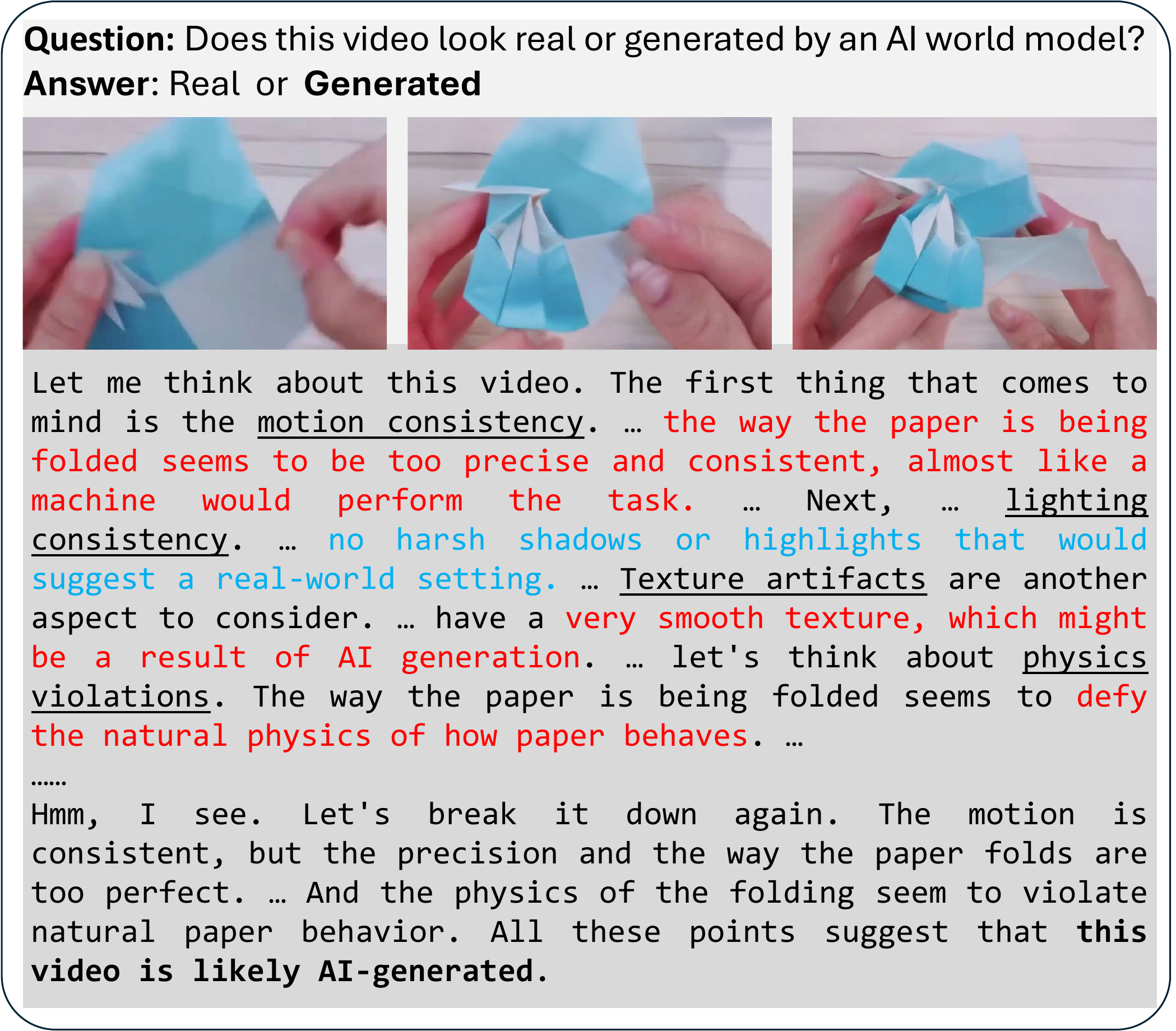}
      \vspace{-0.3in}
    \caption{VidGuard-R1 (GRPO): reasoning about an origami folding sequence}
    \label{fig:ex2}
\end{minipage}
\hfill
\begin{minipage}[t]{0.49\textwidth}
    \centering
    \includegraphics[width=\linewidth]{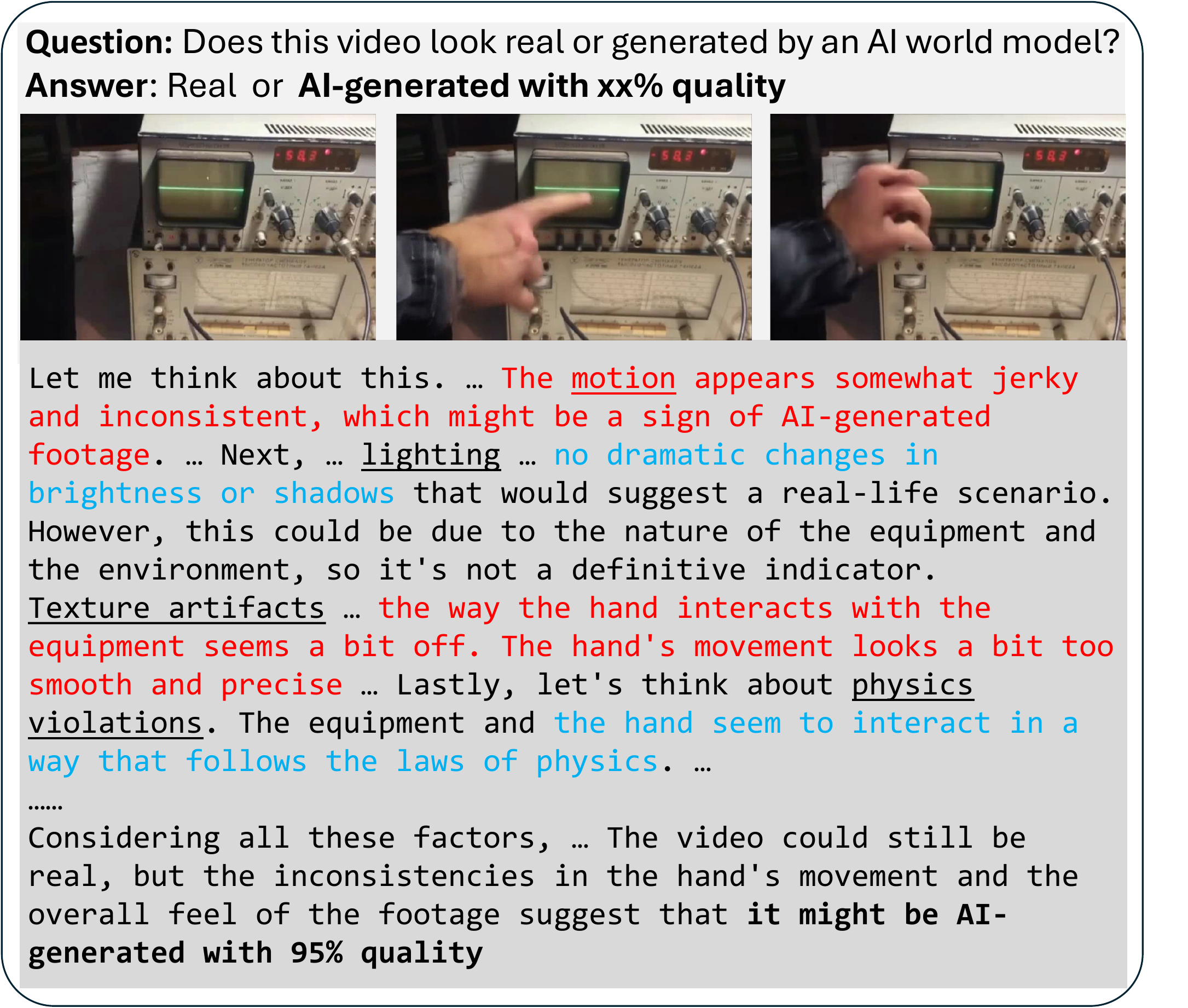}
      \vspace{-0.3in}
    \caption{VidGuard-R1 (GRPO-Q): temporal inconsistency observed as unnatural movements}
    \label{fig:ex4}
\end{minipage}
\end{figure*}

Figures~\ref{fig:ex2} and~\ref{fig:ex4} illustrate cases where \proj{} correctly identifies videos as generated. The model performs multi-faceted reasoning across motion, lighting, texture, and physical plausibility before arriving at a final decision. Notably, it does not rely on a single cue, but instead accumulates evidence across frames, resembling how humans distinguish fake videos. In each figure, pink highlights denote cues suggesting realism, red indicates artifacts indicative of generation, yellow marks intermediate reasoning steps, and underlines represent several key factors.

For instance, in Figure~\ref{fig:ex2}, the smooth hand motion initially suggests realism; however, once the origami folds in a physically implausible manner, the model revises its judgment. In Figure~\ref{fig:ex4}, although the lighting and shadows are consistent—typically a cue for authenticity—the model recognizes that this is insufficient in a largely static scene with only a stationary machine and a human hand. In particular, even in its final prediction, the model reflects on earlier realistic cues and acknowledges that \textit{the video could still be real}, underscoring its nuanced, human-like reasoning in assessing video quality. Additional case studies are provided in the Appendix~\ref{sec:case}.

\vspace{-0.5em}
\section{Conclusion}\label{sec:conclusion}
\vspace{-0.5em}
We propose \proj{}, an MLLM-based discriminator that not only detects AI-generated videos with high accuracy but also provides interpretable reasoning. By leveraging reinforcement learning with reward models targeting temporal artifacts and generation quality, \proj{} achieves 85\% accuracy on our dataset, 97\% on GenVidBench, and 96\% on GenVideo, substantially surpassing prior state-of-the-art methods. We expect this work to advance MLLM-based video analysis and foster future research on strengthening MLLMs’ reasoning.
\vspace{-0.5em}
\subsection{Limitations}
\vspace{-0.5em}

% Our dataset currently includes fake videos generated using HunyuanVideo and CogVideoX, which are the primary open-source models supporting large-scale text--image joint conditioning. Other diffusion models provide only text- or image-based conditioning, limiting their suitability for our pairwise real--fake construction. While the current design ensures strong contextual alignment between real and generated videos, incorporating outputs from a broader set of generative models would further improve diversity and robustness, thereby enhancing applicability to real-world scenarios.

Our dataset currently contains fake videos generated by HunyuanVideo and CogVideoX, the main open source models that support large-scale text-image joint conditioning. Most other diffusion models offer only text or image conditioning, making them less suitable for our pairwise data construction. Although this design ensures strong contextual alignment, adding outputs from more generative models would increase diversity and robustness, improving real-world applicability.
% \newpage
\section*{Ethics Statement}

This work does not involve personally identifiable information or sensitive user data. All datasets used in our experiments are publicly available and constructed in accordance with their licenses and usage guidelines. VidGuard-R1 is designed to mitigate societal risks associated with AI-generated videos, such as misinformation and reputational harm, by providing interpretable CoT reasoning to assist human verification. To the best of our knowledge, the method does not introduce risks related to fairness, safety, or privacy.

\section*{Acknowledgement}

This work was supported in part by the National Science Foundation under Grant CNS-2212297. We are grateful to the anonymous reviewers for their insightful suggestions, which helped improve the quality and clarity of this paper.
% \newpage

\bibliography{iclr2026_conference}
\bibliographystyle{iclr2026_conference}

\clearpage
\appendix
% % \input{sections/related_work}
\section{Additional Setup}\label{sec:ablation}

To further guide the model during RL training, we incorporate a length-based reward strategy. We promote informative yet concise reasoning by rewarding outputs that are neither too brief nor excessively long. Specifically, if the model predicts the correct answer and the length of the response falls within the range \([l_{\min}, l_{\max}]\), an additional reward \(\omega\) is assigned. Let \(l_i\) be the length of the model's response for the \(i\)-th video. The reward is defined as:
\begin{equation}
r_i^{total} = 
\begin{cases}
r_i + \omega, & \text{if } o_i \text{ is correct and } l_{\min} \leq l_i \leq l_{\max} \\
r_i, & \text{otherwise}
\end{cases}
\end{equation}
where we set $\omega = 0.1$, $l_{\min} = 320$, and $l_{\max} = 512$.

\section{Comprehensive Benchmark Evaluation}\label{sec:add_eval}
In this section, we provide extended benchmark results for \proj{} alongside additional MLLMs. Table~\ref{tab:add_genvidbench} presents mean Top-1 accuracy on GenVidBench across multiple video datasets, including CNN and Transformer baselines as well as selected MLLM variants. Table~\ref{tab:add_genvideo} reports comprehensive F1 and recall scores on the GenVideo dataset, including all models provided in the official benchmark alongside our MLLM variants. These extended tables offer a complete comparison of performance across all evaluated models.

\begin{table*}[!t]
\centering
\caption{Extended GenVidBench results with \proj{} and additional MLLMs, reported as mean Top-1 accuracy (\%). TF denotes transformer.}
\resizebox{\textwidth}{!}{%
   \begin{tabular}{ccccccc|c}
   \toprule
   \textbf{Method} & \textbf{Type} & \textbf{MuseV} & \textbf{SVD} & \textbf{CogVideo} & \textbf{Mora} & \textbf{HD-VG} & \textbf{Mean} \\
   \hline
   SlowFast~\citep{slowfast} & CNN& 12.25  & 12.68  & 38.34  & 45.93  & 93.63  & 41.66  \\
   F3Net~\citep{qian2020thinking}  & CNN& 37.43  & 37.27  & 36.46  & 39.59  & 52.76  & 42.52  \\
   I3D~\citep{i3d}& CNN& 8.15  & 8.29  & 60.11  & 59.24  & 93.99  & 49.23  \\
   CFV2~\citep{cfv2} & CNN& 86.26  & 86.53  & 10.10  & 16.90  & 88.40  & 60.53  \\
   TPN~\citep{tpn}& CNN& 37.86  & 8.79  & 68.25  & 90.04  & 97.34  & 61.52  \\
   TIN~\citep{tin}  & CNN& 33.78  & 21.47  & 81.59  & 79.44  & 97.88  & 63.97  \\
   TRN~\citep{trn} & CNN& 38.92  & 26.64  & 91.34  & 93.98  & 93.97  & 71.26  \\
   TSM~\citep{tsm} & CNN& 70.37  & 54.70  & 78.46  & 70.37  & 96.76  & 76.40  \\
   X3D~\citep{x3d} & CNN& 92.39 & 37.27  & 65.72  & 49.60  & 97.51  & 77.09  \\
   \hline 
   UniFormer V2~\citep{uniformerv2} & TF & 20.05  & 14.81  & 45.21  & 99.21 & 96.89  & 57.55  \\
   TimeSformer~\citep{timesformer} & TF & 73.14  & 20.17  & 74.80  & 39.40  & 92.32  & 64.28  \\
   VideoSwin~\citep{videoswin} & TF & 62.29  & 8.01  & 91.82 & 45.83  & \textbf{99.29} & 67.27  \\
   MViT V2~\citep{mvit} & TF & 76.34  & \textbf{98.29} & 47.50  & 96.62  & 97.58  & 79.90 \\
   \hline 
   \midrule
   Qwen2.5-VL-7B~\citep{Qwen2.5-VL} & MLLM & 25.86  &  27.06  & 68.51  & 43.26  & 71.15 & 47.30 \\
   GPT-4.1 mini~\citep{gpt4} & MLLM & 26.07  & 33.78 & 94.07  & 57.19  & 87.64  & 59.62 \\
   VidGuard-R1 (CoT) & MLLM& 36.52  & 16.02 & 99.35  & 76.94  & 99.94  &66.09 \\
   \makecell[c]{VidGuard-R1 (GRPO,\\ GenVideo-pretrained, Zero-shot)}
   & MLLM&  97.24 & 96.59 & 99.88  & 99.93  & 88.14  & 96.37 \\
   VidGuard-R1 (GRPO) & MLLM & \textbf{97.38}  & 94.98 & \textbf{99.90}  & \textbf{99.99}  & 95.46  & \textbf{97.53} \\
   \hline
   \end{tabular}%
}
\label{tab:add_genvidbench}
\end{table*}

\begin{table*}[!t]
    \centering
    \caption{Extended GenVideo results with \proj{} and additional MLLMs, evaluated by F1 and recall scores}
    \label{tab:add_genvideo}
    \resizebox{\textwidth}{!}{
    \begin{tabular}{ccccccccccccc|c}
    \toprule
       \multirow{2}{*}{Model} & Detection&\multirow{2}{*}{Metric}&\multirow{2}{*}{Sora}&Morph&\multirow{2}{*}{Gen2}&\multirow{2}{*}{HotShot}&\multirow{2}{*}{Lavie}&\multirow{2}{*}{Show-1}&Moon&\multirow{2}{*}{Crafter}&Model&Wild&\multirow{2}{*}{Mean}\\
       &level&&&Studio&&&&&Valley&&Scope&Scrape\\
       \midrule
       \multirow{2}{*}{F3Net~\citep{qian2020thinking}}&\multirow{2}{*}{Image}&
       R& 
       0.83& 0.99 &0.98
       &0.77&0.57 &0.36 &0.99&0.99
       & 0.89& 0.76&0.81 
       \\
       &&F1&0.50&0.94&0.96&0.81&0.69&0.49&
0.93&0.96&0.88&0.82&0.80
        \\
       
        \midrule
       \multirow{2}{*}{NPR~\citep{npr}}&\multirow{2}{*}{Image}&
R& 0.91 &0.99& 0.99& 0.24& 0.89& 0.57 &0.97
       &0.99&0.94&0.87&0.84\\
       &&F1&0.27&0.84&0.91&0.30&0.86&0.59&
0.81&0.91&0.81&0.81&0.71
       \\
       
       \midrule
       \multirow{2}{*}{STIL~\citep{stil}}&\multirow{2}{*}{Video}
        &R
       & 0.78 & 0.98 & 0.98 & 0.76 & 0.61 & 0.53 & 0.99 & 0.97 & 0.94 & 0.65 & 0.82
       \\
       &&F1&0.38&0.90&0.94&0.78&0.72&0.62&0.90&0.94&0.88&0.72&0.78
       \\
       \midrule
       \multirow{2}{*}{VideoMAE~\citep{videomae}}&\multirow{2}{*}{Video}&
        R&
       0.67 & 0.96 &0.98 & 0.96& 0.77 & 0.80 &0.97&0.96& 0.96
       &0.68&0.87\\
        &&F1&0.62&0.95&0.98&\textbf{0.96}&0.86&0.87&0.96&0.97&\textbf{0.96}&0.79&0.89
       \\

       \midrule       
\multirow{2}{*}{MINTIME-CLIP~\citep{minetime}}&\multirow{2}{*}{Video}&
R& 0.89&1.00 &0.98 &0.26 &0.96&0.98 &0.99 &  1.00&0.84 &0.82 &0.87
\\
&&F1&0.49 &0.93 & 0.96 &0.37&0.94 &0.92 &0.92 & 0.96& 0.84&0.85 &0.82\\

\midrule
       \multirow{2}{*}{FTCN-CLIP~\citep{ftcn}}&\multirow{2}{*}{Video}&
R&0.87 & 1.00 & 0.98 & 0.17 & 0.97 & 0.91 & 1.00 & 1.00 &  0.85& 0.82& 0.86 \\
&&F1& 0.78 & 0.98 & 0.98 & 0.29 & 0.98 & 0.94 & 0.98 & \textbf{0.99} & 0.90 & 0.89 & 0.87\\

\midrule

       \multirow{2}{*}{TALL~\citep{tall}}&\multirow{2}{*}{Video}&
R&0.91& 0.98& 0.97& 0.83& 0.76& 0.79&0.99 & 0.98 &0.94 &0.66 & 0.88 \\
&&F1&0.26& 0.82& 0.89& 0.74&  0.77 &0.72& 0.81&0.90 & 0.80 & 0.67 &0.74   \\

       \midrule
       \multirow{2}{*}{CLIP~\citep{clip}}&\multirow{2}{*}{Image}&
 R&
       0.94 & 0.99& 0.91& 0.77& 0.88& 0.86& 0.99& 0.99& 0.84&0.84&0.90\\
       &&F1&0.28&0.84&0.86&0.72&0.85&0.76&0.82&0.91&0.76&0.79&0.76
       \\

       \midrule
        \multirow{2}{*}{DeMamba-CLIP~\citep{demamba}}&\multirow{2}{*}{Video}&R& 0.95 & 1.00 & 0.98 & 0.69 & 0.92 & 0.93 & 1.00 & 1.00 & 0.83 & 0.82 &0.91\\
        
        &&F1& 0.64&0.96&0.97&0.78&0.94&0.92&0.95&0.98&0.87&0.87&0.89
       \\

        \midrule

       \multirow{2}{*}{XCLIP~\citep{xclip}}&\multirow{2}{*}{Video}&
        R&
       0.82&0.99 &0.93 & 0.61 & 0.79&0.69 &0.97&0.99& 0.77 & 0.83&0.84\\
       &&F1&0.31&0.88&0.90&0.65&0.82&0.70&0.86&0.93&0.75&0.82&0.76\\
       \midrule
       \multirow{2}{*}{DeMamba-XCLIP~\citep{demamba}}&\multirow{2}{*}{Video}&
        R
       &0.98& 1.00 & 0.99 & 0.65 & 0.94 & 0.98 & 1.00& 1.00&0.92 &0.89 & 0.93\\
       &&F1&0.64& 0.96& 0.97& 0.75& \textbf{0.95}&0.95&0.95&0.97&0.92&\textbf{0.91}&0.90
        \\
        \midrule
       \midrule
        \multirow{2}{*}{Qwen2.5-VL-7B~\citep{Qwen2.5-VL}}&\multirow{2}{*}{MLLM}&
        R&
       0.58 &0.56 &0.54 & 0.33 & 0.43 &0.38 &0.81 &0.63 & 0.51 & 0.70 &0.54 \\
       &&F1&0.74 &0.72 &0.70 &0.49 &0.60 &0.55 &0.90 &0.77 &0.68 &0.82 &0.70 \\
       \midrule
        \multirow{2}{*}{GPT-4.1 mini~\citep{gpt4}}&\multirow{2}{*}{MLLM}&
        R&
       0.43 &0.67  &0.56  & 0.54  & 0.63 &0.56  &0.92 &0.67 & 0.69  & 0.69 &0.65 \\
       &&F1&0.60 &0.80 &0.72&0.70&0.77 &0.72 &0.96 &0.80 &0.82 &0.82 &0.72 \\
       \midrule
        \multirow{2}{*}{VidGuard-R1 (CoT)}&\multirow{2}{*}{MLLM}& 
        R&
       0.92 & 0.89 & 0.91 & 0.90 & 0.98 & 0.79 &0.99 &0.85  & 0.89  & 0.87 &0.90 \\
       &&F1&0.90 &0.91 &0.95 &0.89 &0.99 &0.81 & 0.95 & 0.89& 0.85&0.88 & 0.90 \\
       \midrule
        \multirow{2}{*}{\makecell[c]{VidGuard-R1 (GRPO,\\ GenVidBench-pretrained, Zero-shot)}
}&\multirow{2}{*}{MLLM}&
        R&
       0.95 &0.98 &0.90 & 0.89 & 0.97 & 0.85 & 0.99 & 0.93 & 0.81 & 0.87 & 0.92\\
       && F1 & 0.93& 0.93 & 0.96 & 0.91 & 0.99 & 0.82 & 0.95 & 0.89& 0.85& 0.88 & 0.91\\

       \midrule
        \multirow{2}{*}{VidGuard-R1 (GRPO)}&\multirow{2}{*}{MLLM}&
        R&
       0.95&1.00 &0.98 & 0.94& 0.98 & 0.95 & 0.97 & 0.99 & 0.94 & 0.91 & 0.96\\
       && F1 & \textbf{0.97} & \textbf{0.99} & \textbf{0.99} &0.91 & \textbf{0.99} & 0.89 & \textbf{0.99} & \textbf{0.99} & 0.95& 0.90& \textbf{0.96}\\

        \bottomrule
    \end{tabular}}

\end{table*}

\section{Zero-Shot Generalization to Unseen Generative Models}
To evaluate the robustness of \proj{} beyond the curated training sources, we assess its zero-shot performance on a diverse set of recently released generative video models that were not used during training, including Gen-3 Alpha \citep{runway2025}, Pika \citep{pika2025}, Pika~2.2 \citep{pika2025}, Luma Ray2 \citep{luma2025}, Sora \citep{sora2025}, Veo2 \citep{veo2025}, Veo3 \citep{veo2025}, and Wan~2.1 \citep{wan2025}.  
Table~\ref{tab:zeroshot} summarizes performance across these unseen systems. VidGuard-R1 achieves accuracy above 80\% in all cases, reaching up to 96.36\%, demonstrating strong generalization to more recent and increasingly realistic generative models.

\begin{table}[t]
\centering
\caption{{Zero-shot detection accuracy on unseen generative models}}
\label{tab:zeroshot}
\begin{tabular}{l c c c c}
\toprule
\textbf{{Model}} & \textbf{{Total}} & \textbf{{Correct}} & \textbf{{Incorrect}} & \textbf{{Accuracy (\%)}} \\
\midrule
{Gen-3 Alpha}  & {56}  & {49}  & {7}   & {87.50} \\
{Pika}         & {110} & {101} & {9}   & {91.82} \\
{Pika 2.2}     & {110} & {106} & {4}   & {96.36} \\
{Luma Ray2}    & {110} & {98}  & {12}  & {89.09} \\
{Sora}         & {110} & {102} & {8}   & {92.73} \\
{Veo2}         & {52}  & {45}  & {7}   & {86.54} \\
{Veo3}         & {55}  & {45}  & {10}  & {81.82} \\
{Wan2.1}       & {55}  & {46}  & {9}   & {83.64} \\
\bottomrule
\end{tabular}
\end{table}

These analyses demonstrate that \proj{} generalizes effectively beyond the curated generative sources and remains robust across a wide range of unseen, high-quality video generation models.

\section{Prompt}\label{sec:prompt}

\begin{figure*}[!ht]
    \centering
    \includegraphics[width=1\textwidth]{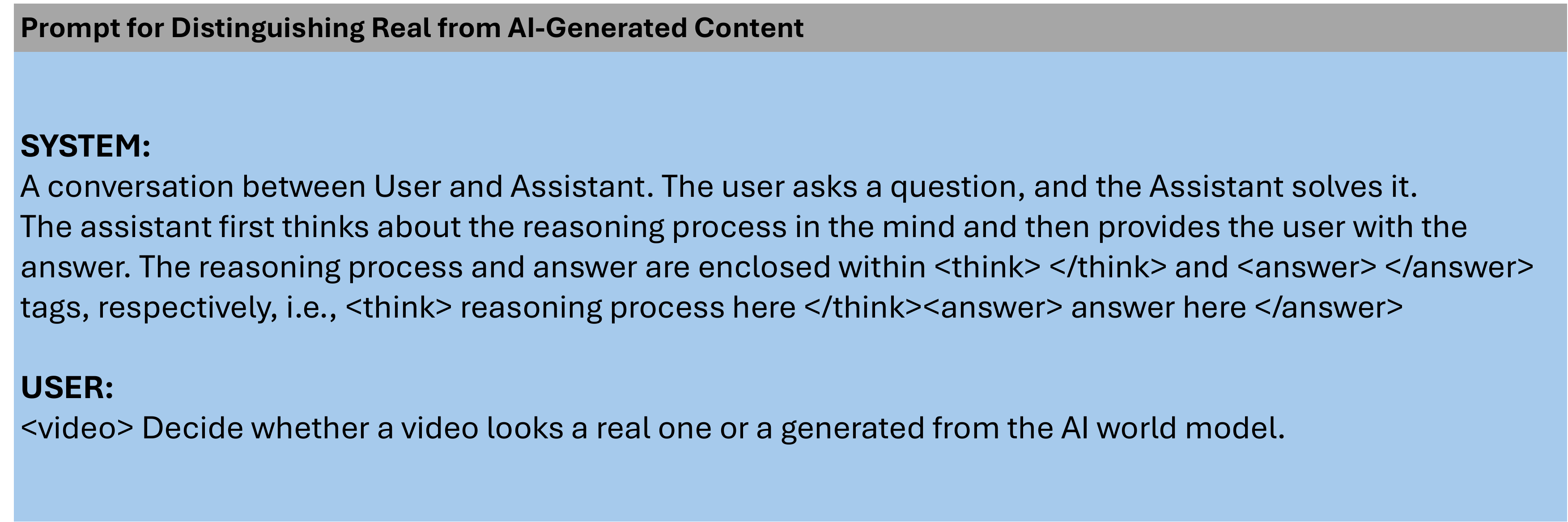}
    \caption{Prompt for identifying realism cues in real videos across four categories}
   \label{fig:prompt_detection}
\end{figure*}

Figure~\ref{fig:prompt_detection} shows the base prompt used for the real-vs-fake classification task. Annotators are instructed to assess whether a video is real or AI-generated by analyzing key visual and physical properties.

\begin{figure*}[!ht]
    \centering
    \includegraphics[width=1\textwidth]{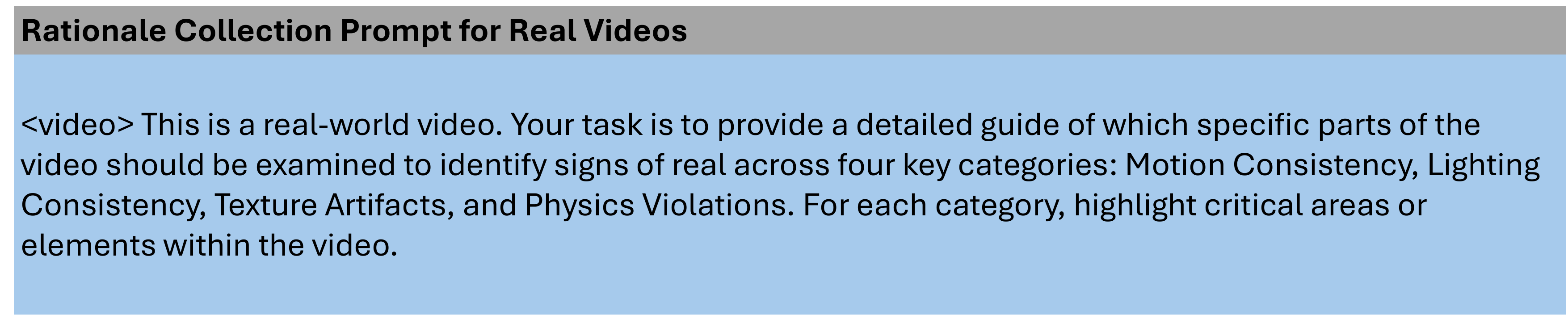}
    \caption{Prompt for identifying realism cues in real videos across four categories}
   \label{fig:prompt_real}
\end{figure*}

\begin{figure*}[!ht]
    \centering
    \includegraphics[width=1\textwidth]{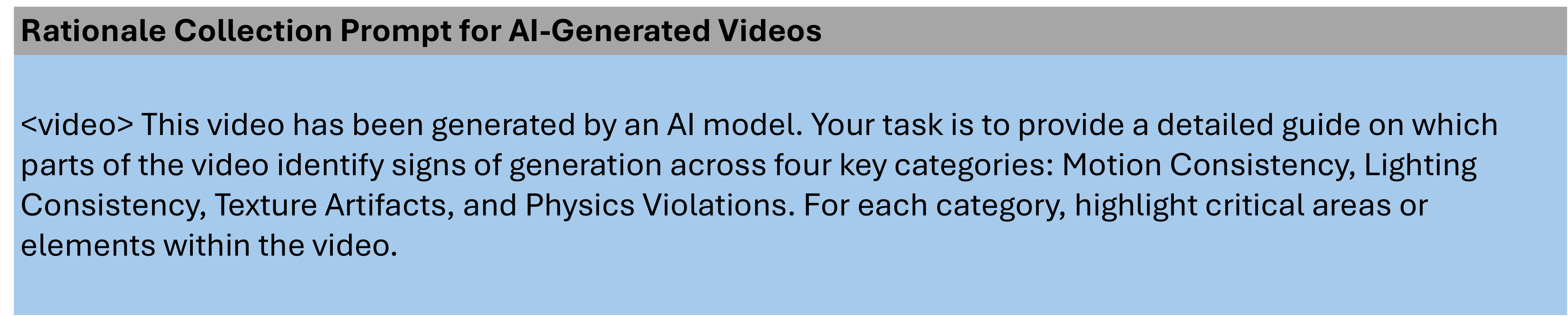}
    \caption{Prompt for identifying artifacts in AI-generated videos across four categories}
   \label{fig:prompt_fake}
\end{figure*}

Figures~\ref{fig:prompt_real} and~\ref{fig:prompt_fake} provide category-specific rationale collection prompts. In particular, Figure~\ref{fig:prompt_real} presents the prompt for identifying visual cues of realism in real videos, while Figure~\ref{fig:prompt_fake} focuses on spotting artifacts in AI-generated videos. Both prompts guide annotators to evaluate videos across four diagnostic categories: motion consistency, lighting consistency, texture artifacts, and physics violations.

\begin{figure*}[!ht]
    \centering
    \includegraphics[width=1\textwidth]{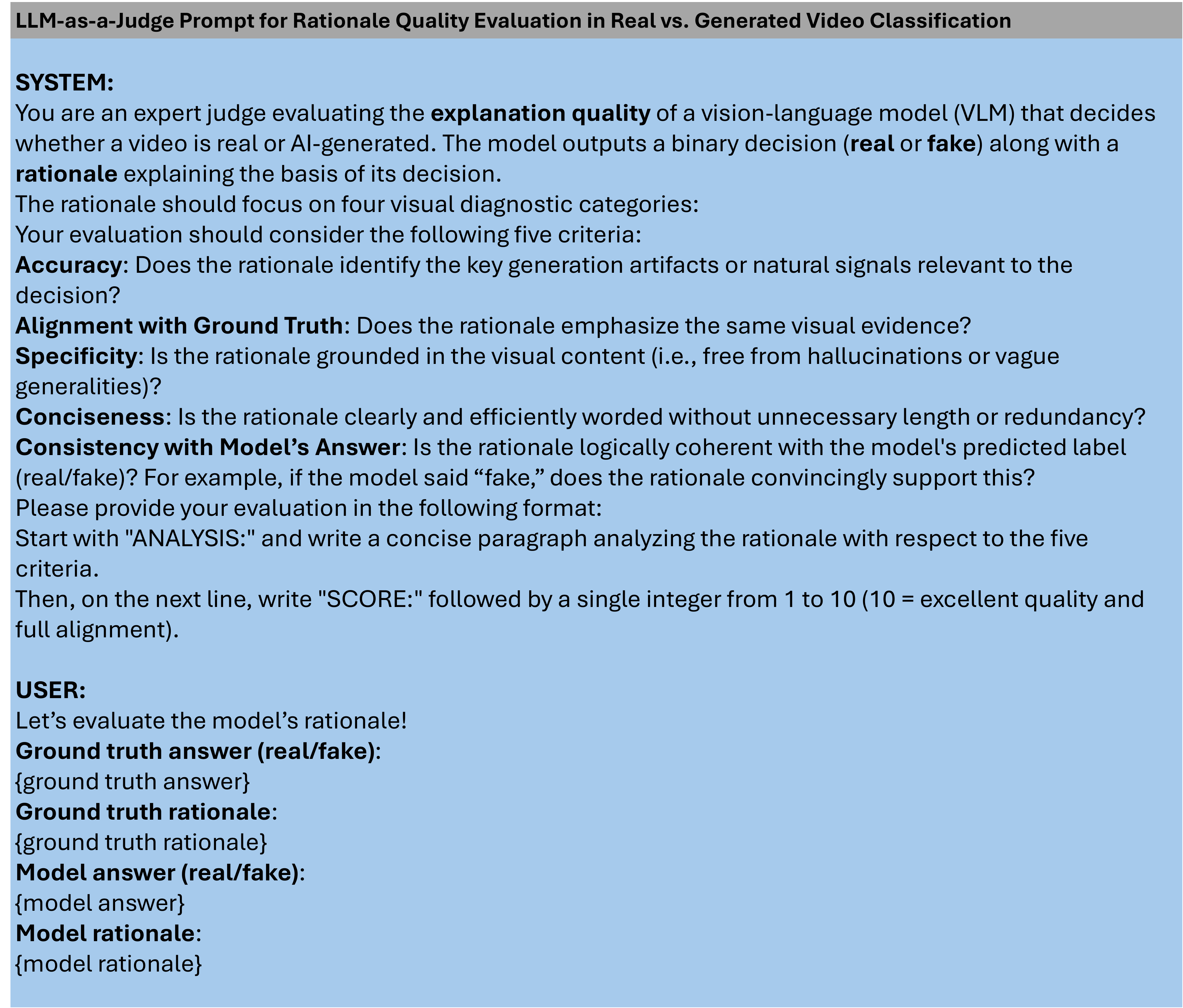}
    \caption{LLM-as-a-judge prompt for rationale quality evaluation}
   \label{fig:prompt_llm_as_a_judge}
\end{figure*}

Figure~\ref{fig:prompt_llm_as_a_judge} illustrates the LLM-as-a-judge prompt used to evaluate rationale quality. In this setting, GPT-4.1 mini rates the quality of model-generated explanations on a 1–10 scale, where a score of 10 corresponds to excellent quality and full alignment with the ground truth rationale.

\section{Case studies on explanations}\label{sec:case}

\subsection{GenVidBench}
\begin{figure*}[!ht]
    \centering
    \includegraphics[width=1\textwidth]{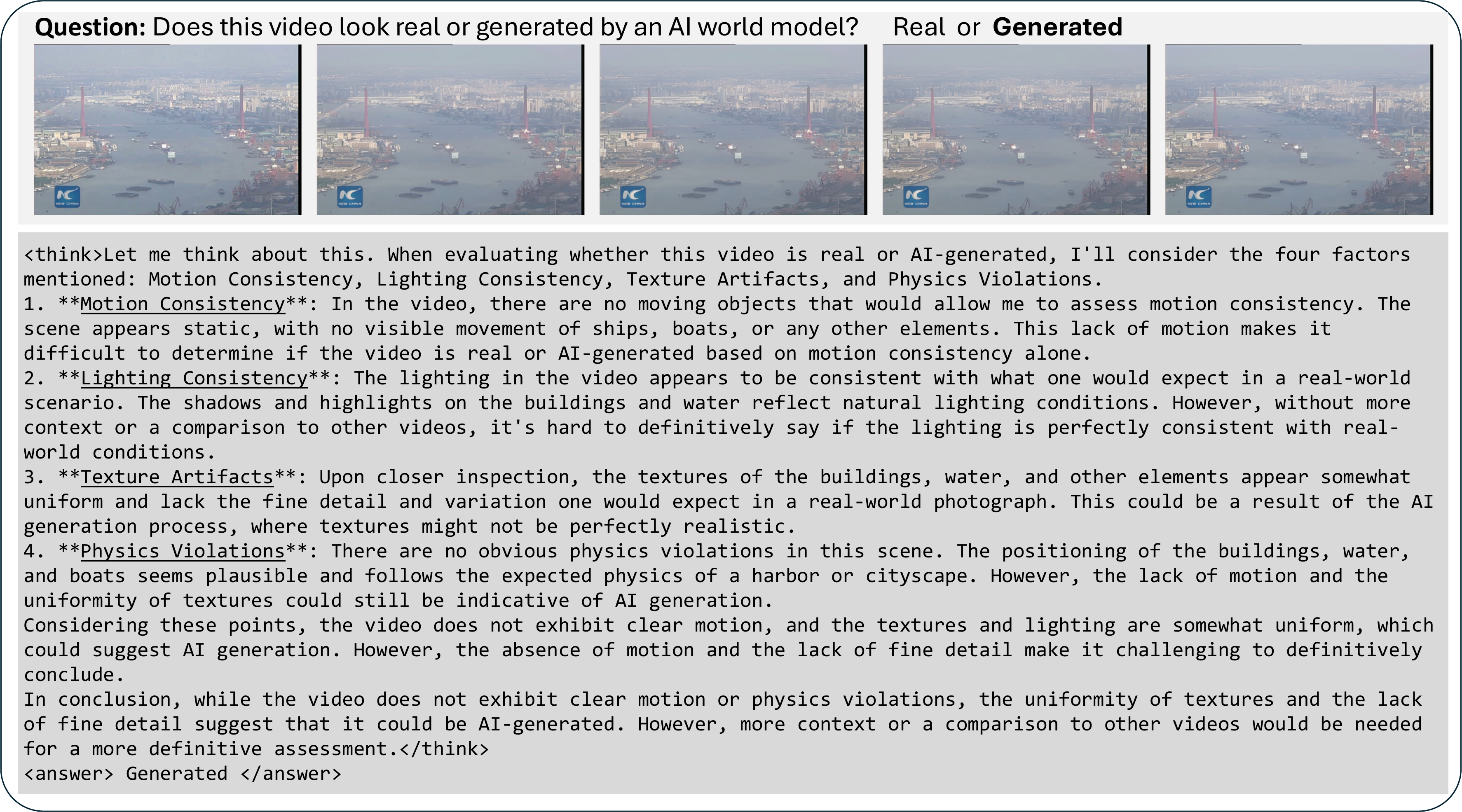}
    \caption{VidGuard-R1 (GRPO) inference on a MuseV-generated video}
   \label{fig:ex_genvidbench_musev}
\end{figure*}

\begin{figure*}[!ht]
    \centering
    \includegraphics[width=1\textwidth]{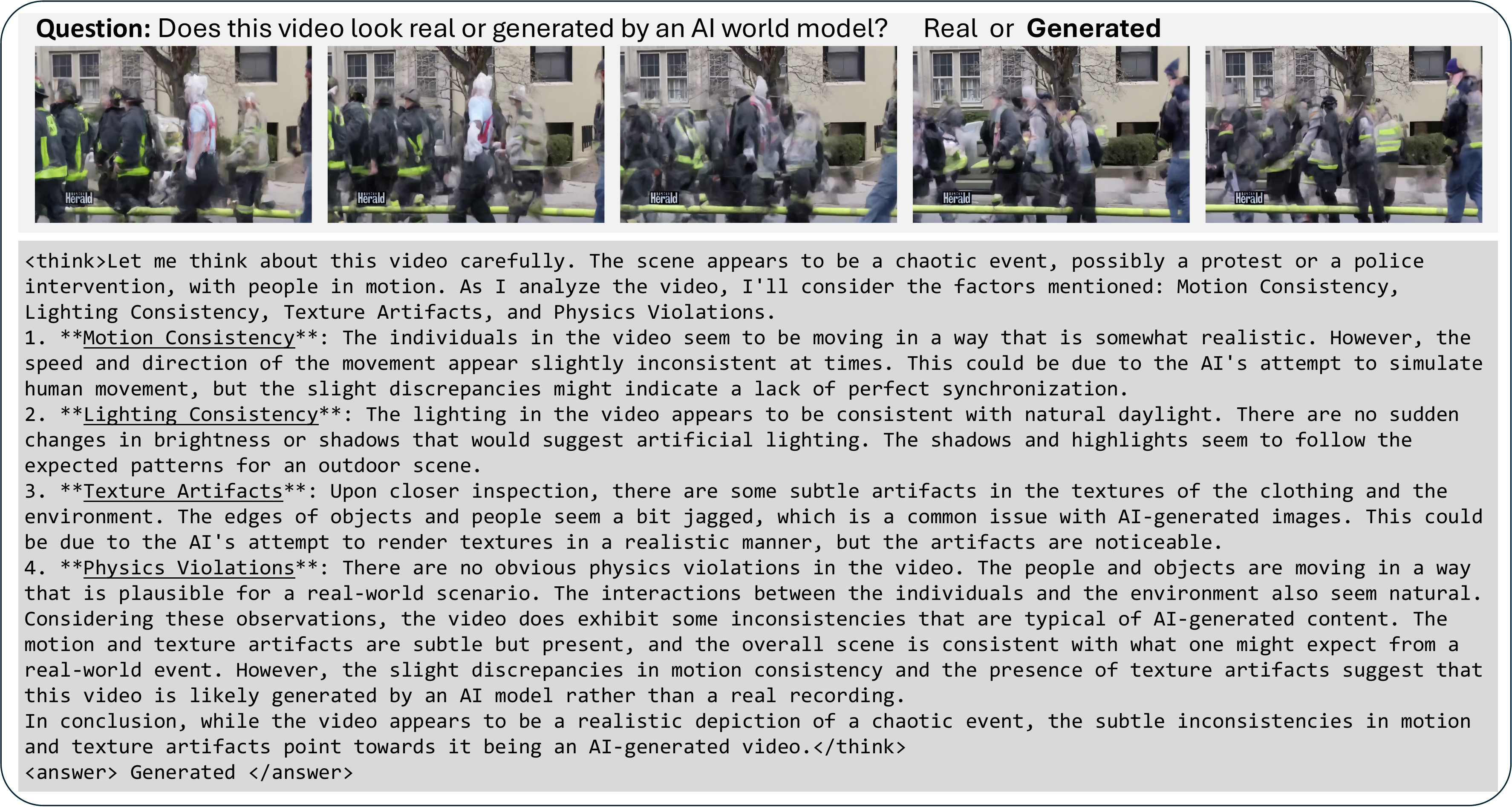}
    \caption{VidGuard-R1 (GRPO) inference on a SVD-generated video}
   \label{fig:ex_genvidbench_svd}
\end{figure*}

\begin{figure*}[!ht]
    \centering
    \includegraphics[width=1\textwidth]{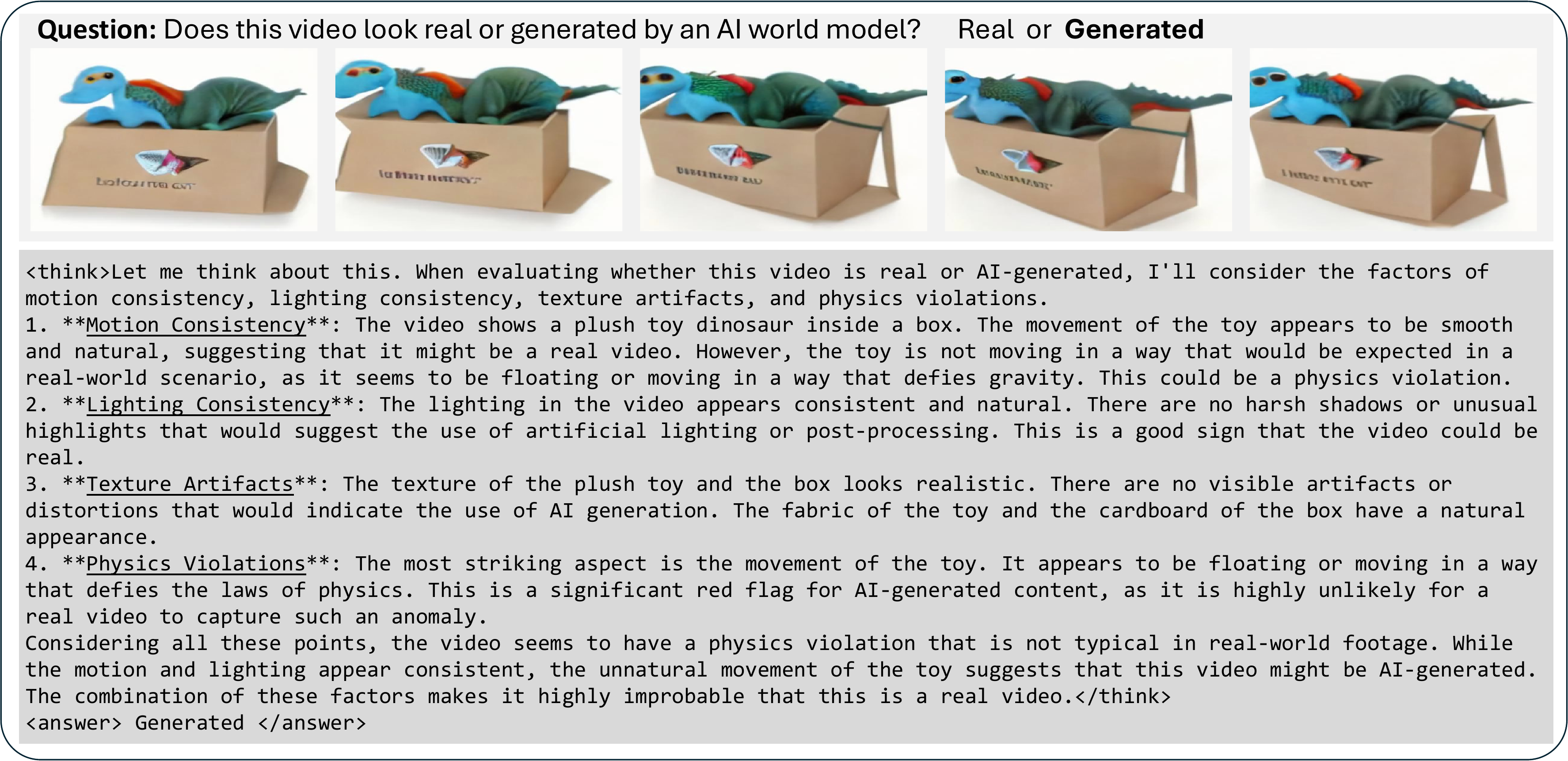}
    \caption{VidGuard-R1 (GRPO) inference on a CogVideo-generated video}
   \label{fig:ex_genvidbench_cogvideo}
\end{figure*}

\begin{figure*}[!ht]
    \centering
    \includegraphics[width=1\textwidth]{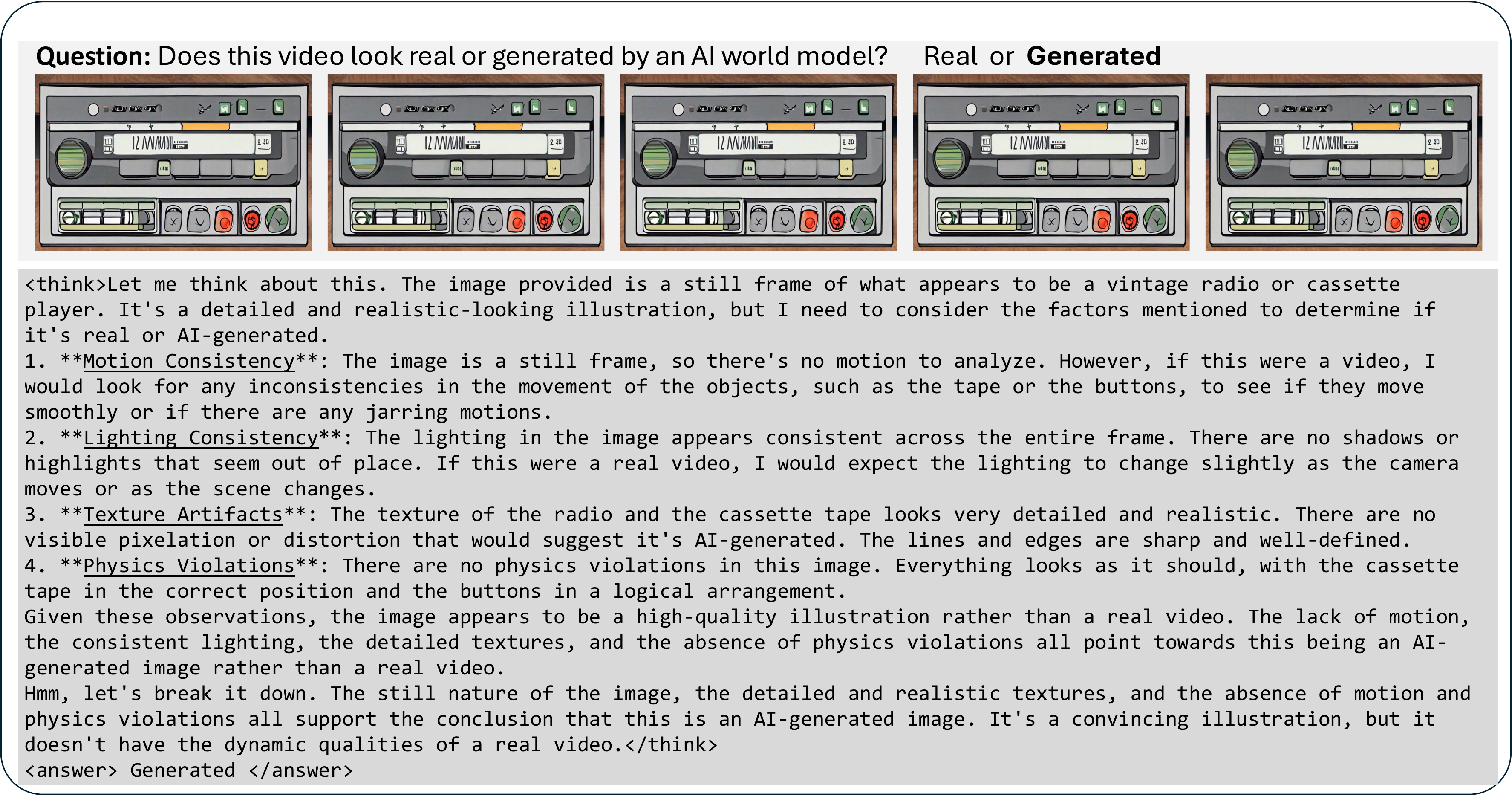}
    \caption{VidGuard-R1 (GRPO) inference on a Mora-generated video}
   \label{fig:ex_genvidbench_mora}
\end{figure*}

Figures~\ref{fig:ex_genvidbench_musev}--\ref{fig:ex_genvidbench_mora} present inference examples for videos synthesized by four distinct AI models included in the GenVidBench testing dataset: MuseV~\citep{musev}, SVD~\citep{svd}, CogVideo~\citep{cogvideo}, and Mora~\citep{mora}.

\subsection{GenVideo}
\begin{figure*}[!ht]
    \centering
    \includegraphics[width=1\textwidth]{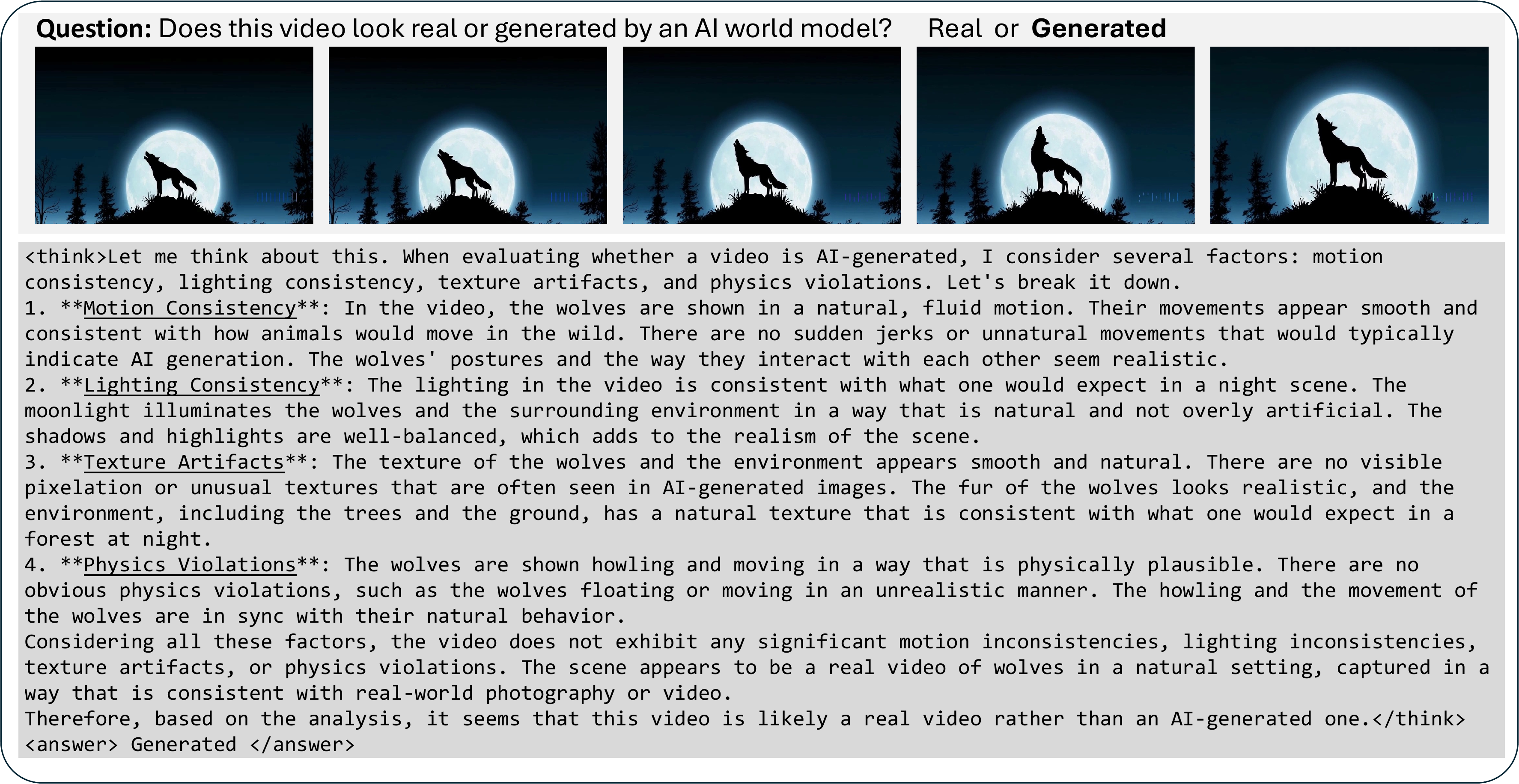}
    \caption{VidGuard-R1 (GRPO) inference on a Sora-generated video}
   \label{fig:ex_genvideo_sora}
\end{figure*}

\begin{figure*}[!ht]
    \centering
    \includegraphics[width=1\textwidth]{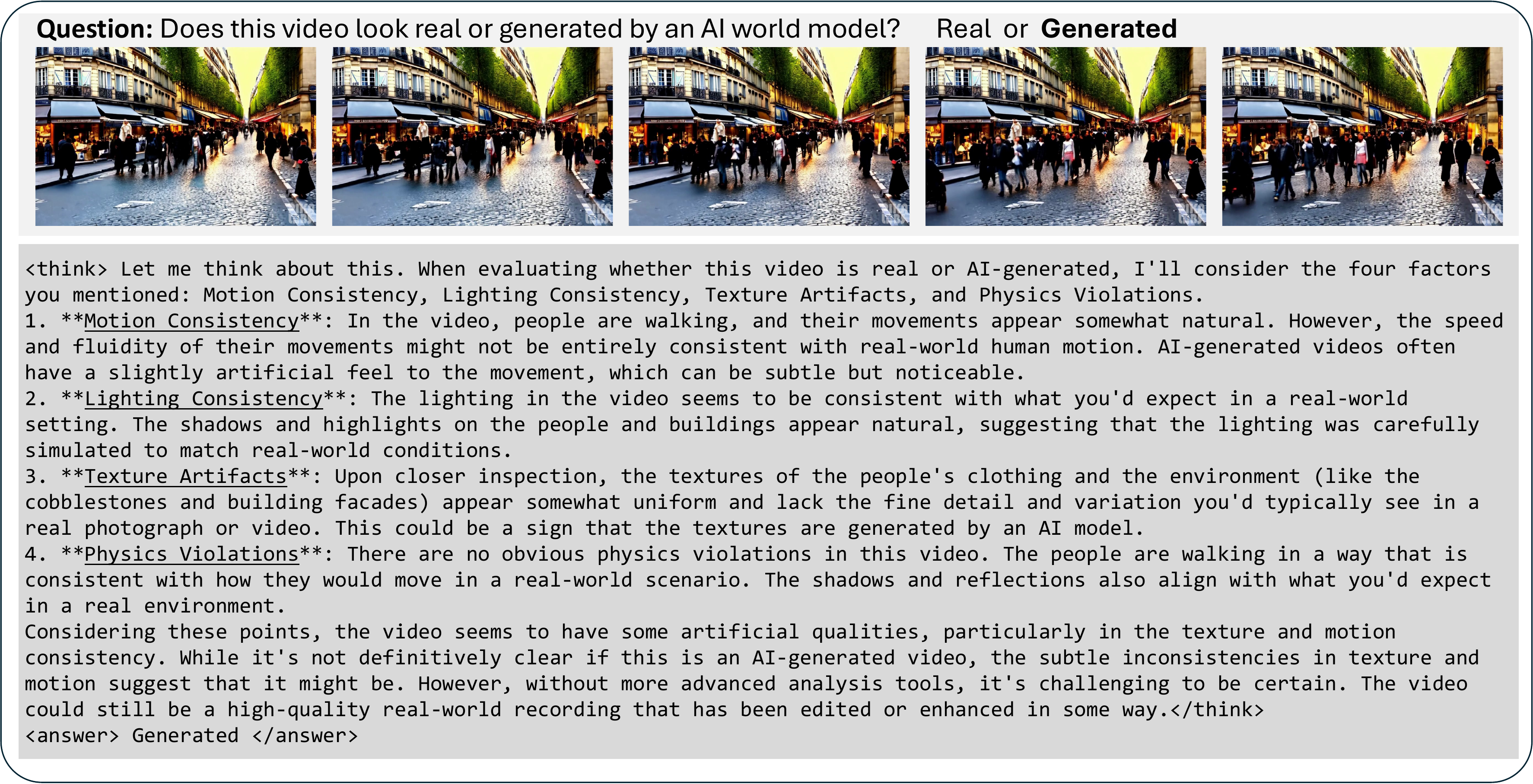}
    \caption{VidGuard-R1 (GRPO) inference on a Morph Studio-generated video}
   \label{fig:ex_genvideo_morphstudio}
\end{figure*}

\begin{figure*}[!ht]
    \centering
    \includegraphics[width=1\textwidth]{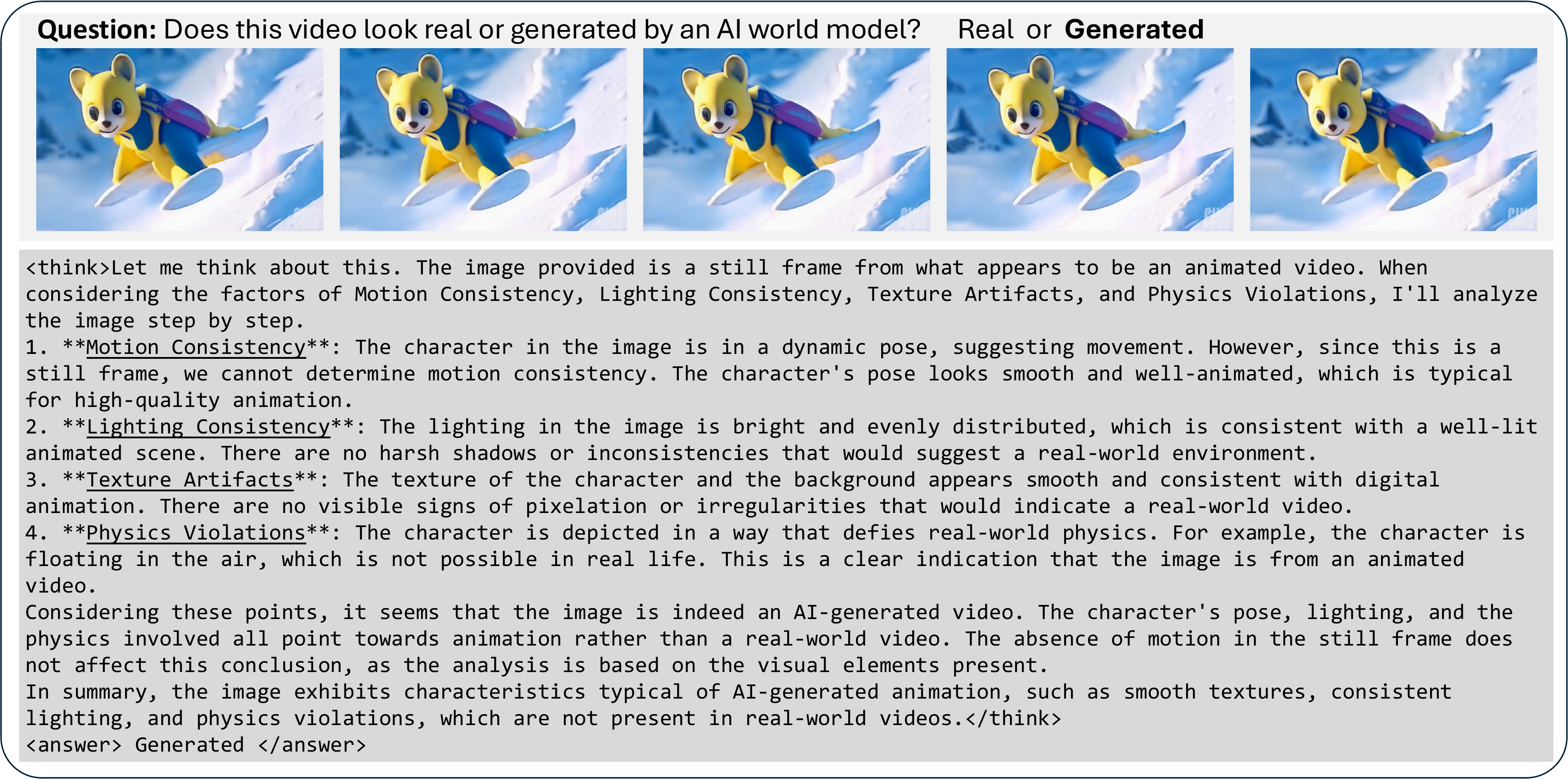}
    \caption{VidGuard-R1 (GRPO) inference on a Gen2-generated video}
   \label{fig:ex_genvideo_gen2}
\end{figure*}

\begin{figure*}[!ht]
    \centering
    \includegraphics[width=1\textwidth]{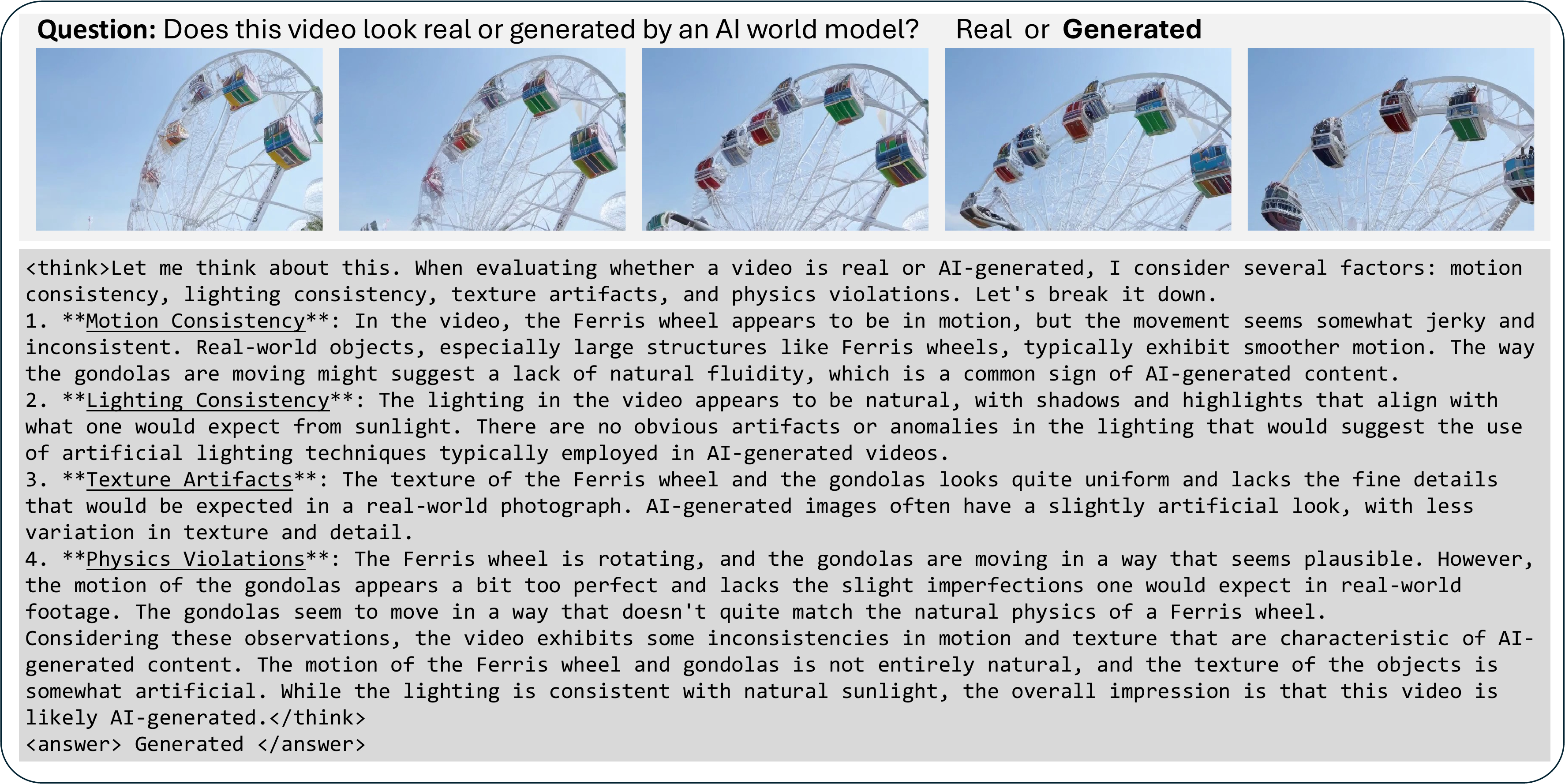}
    \caption{VidGuard-R1 (GRPO) inference on a HotShot-generated video}
   \label{fig:ex_genvideo_hotshot}
\end{figure*}

\begin{figure*}[!ht]
    \centering
    \includegraphics[width=1\textwidth]{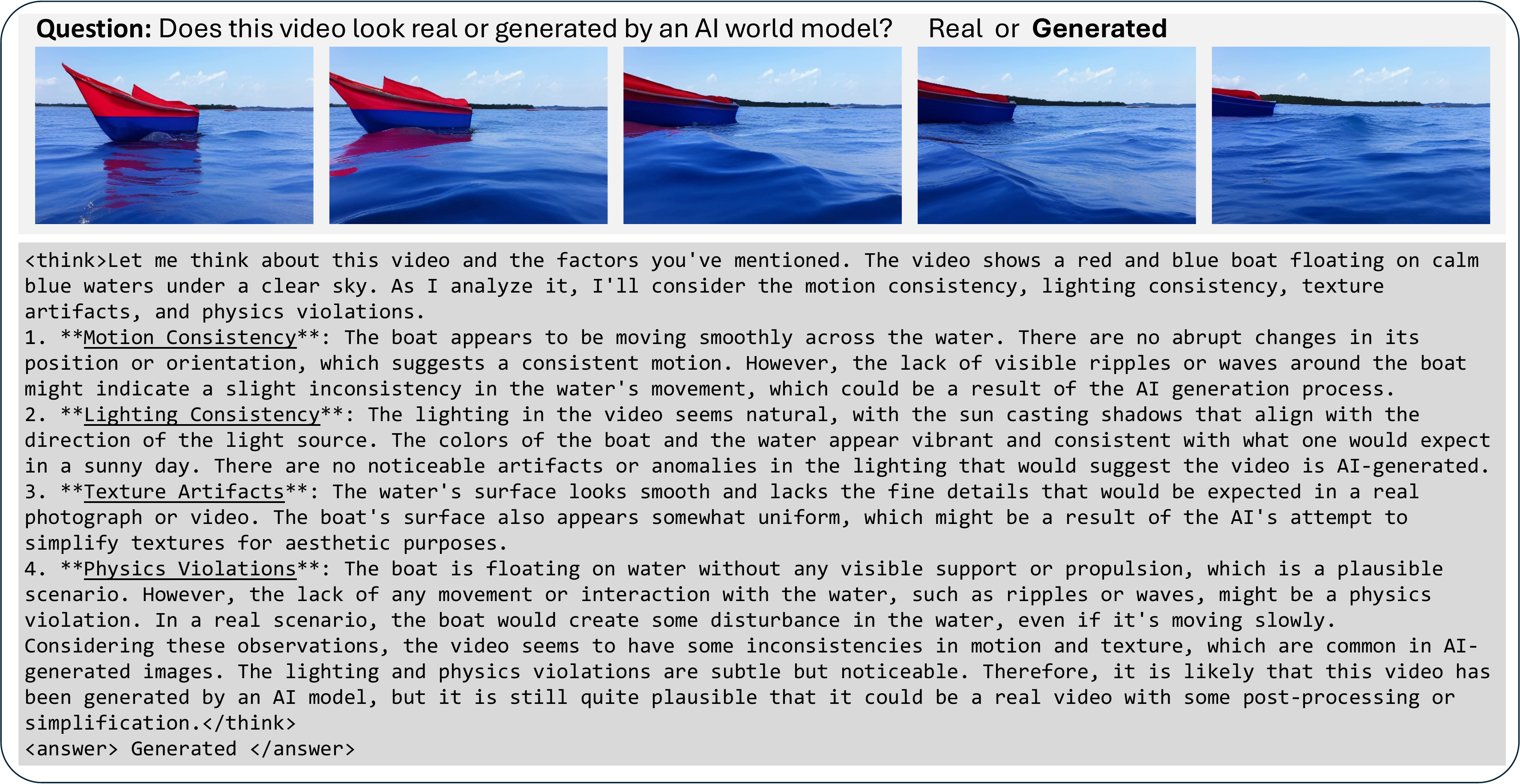}
    \caption{VidGuard-R1 (GRPO) inference on a Lavie-generated video}
   \label{fig:ex_genvideo_lavie}
\end{figure*}

\begin{figure*}[!ht]
    \centering
    \includegraphics[width=1\textwidth]{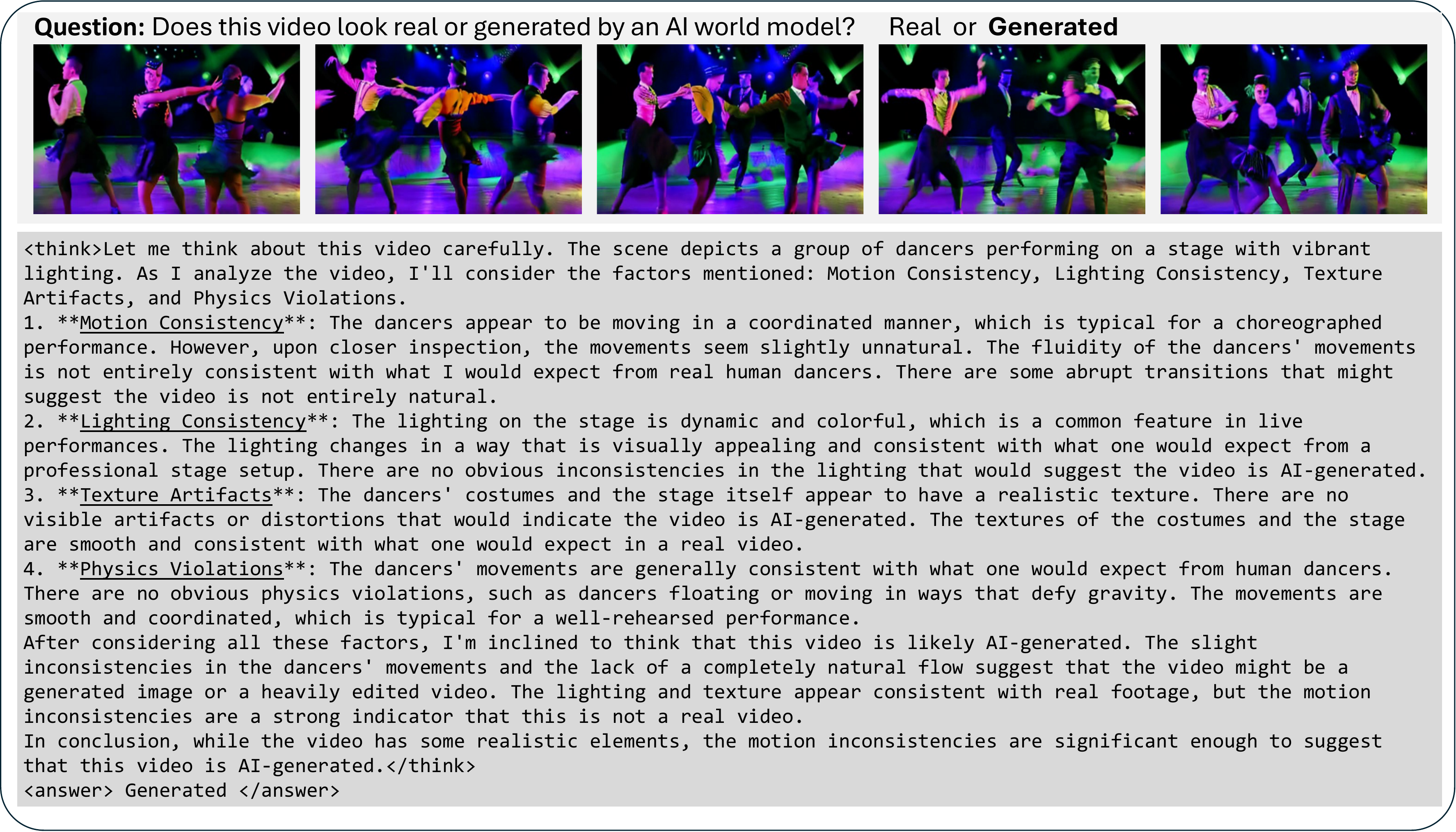}
    \caption{VidGuard-R1 (GRPO) inference on a Show-1-generated video}
   \label{fig:ex_genvideo_show1}
\end{figure*}

\begin{figure*}[!ht]
    \centering
    \includegraphics[width=1\textwidth]{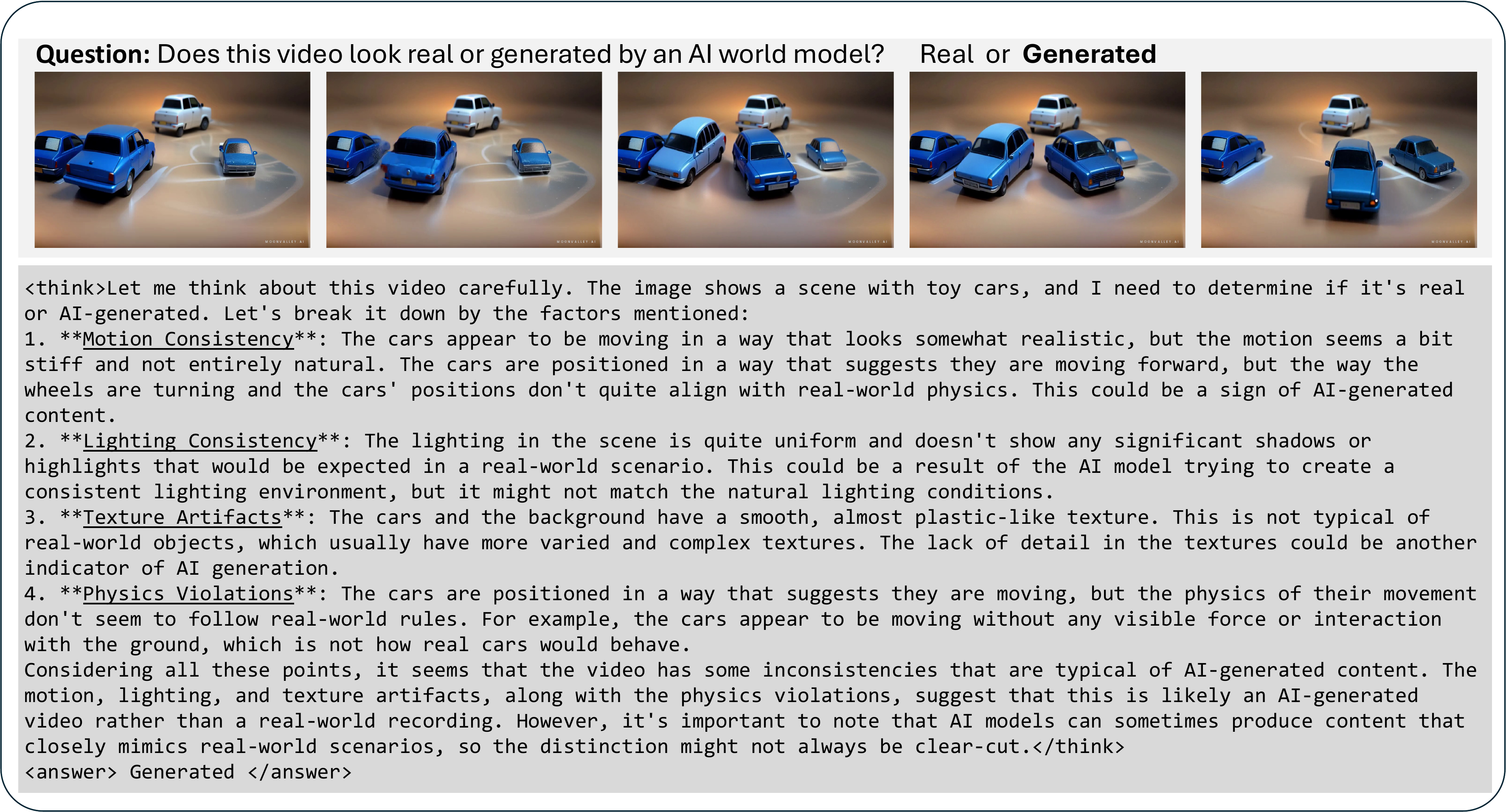}
    \caption{VidGuard-R1 (GRPO) inference on a Moonvalley-generated video}
   \label{fig:ex_genvideo_moonvalley}
\end{figure*}

\begin{figure*}[!ht]
    \centering
    \includegraphics[width=1\textwidth]{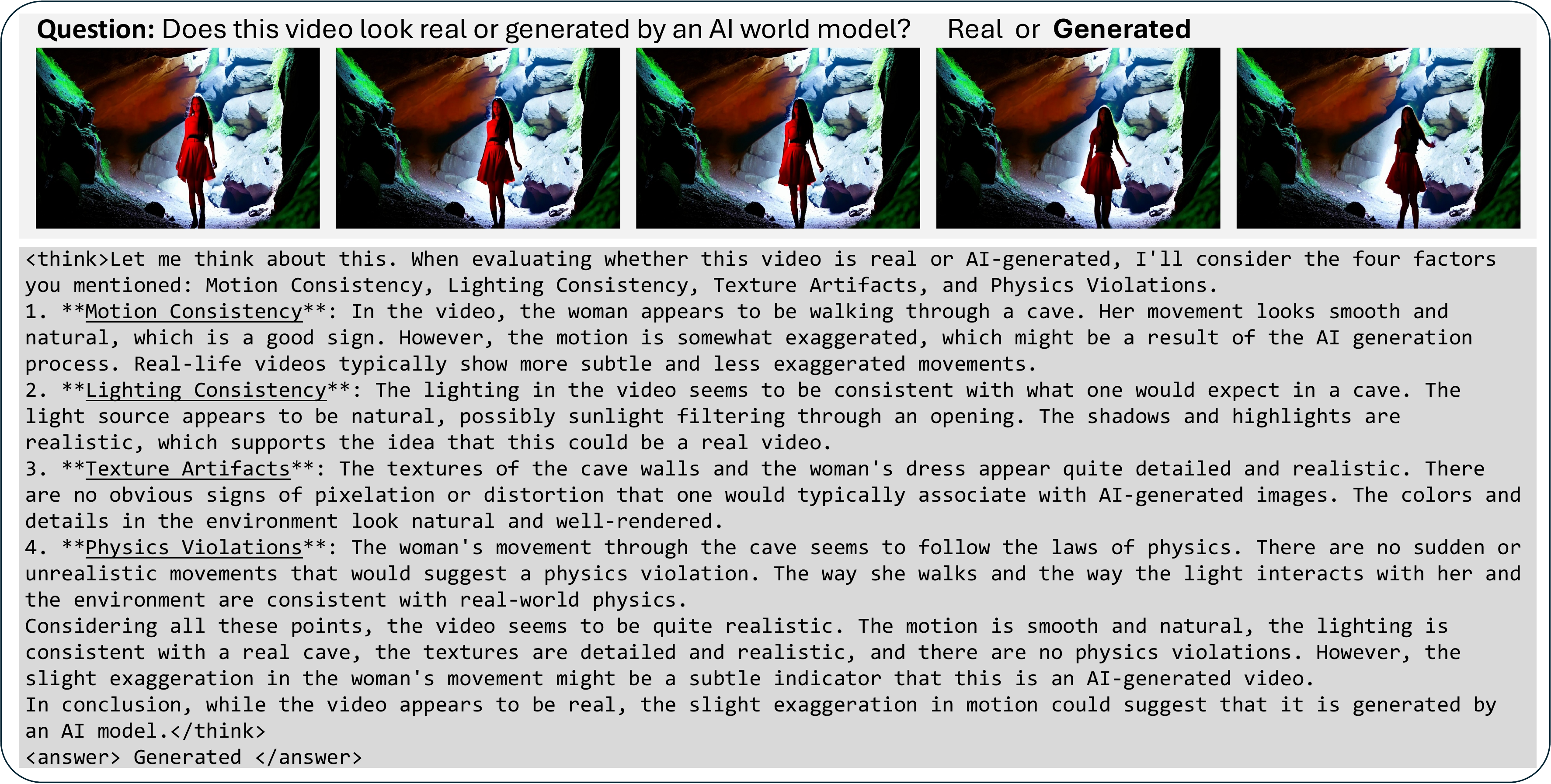}
    \caption{VidGuard-R1 (GRPO) inference on a Crafter-generated video}
   \label{fig:ex_genvideo_crafter}
\end{figure*}

\begin{figure*}[!ht]
    \centering
    \includegraphics[width=1\textwidth]{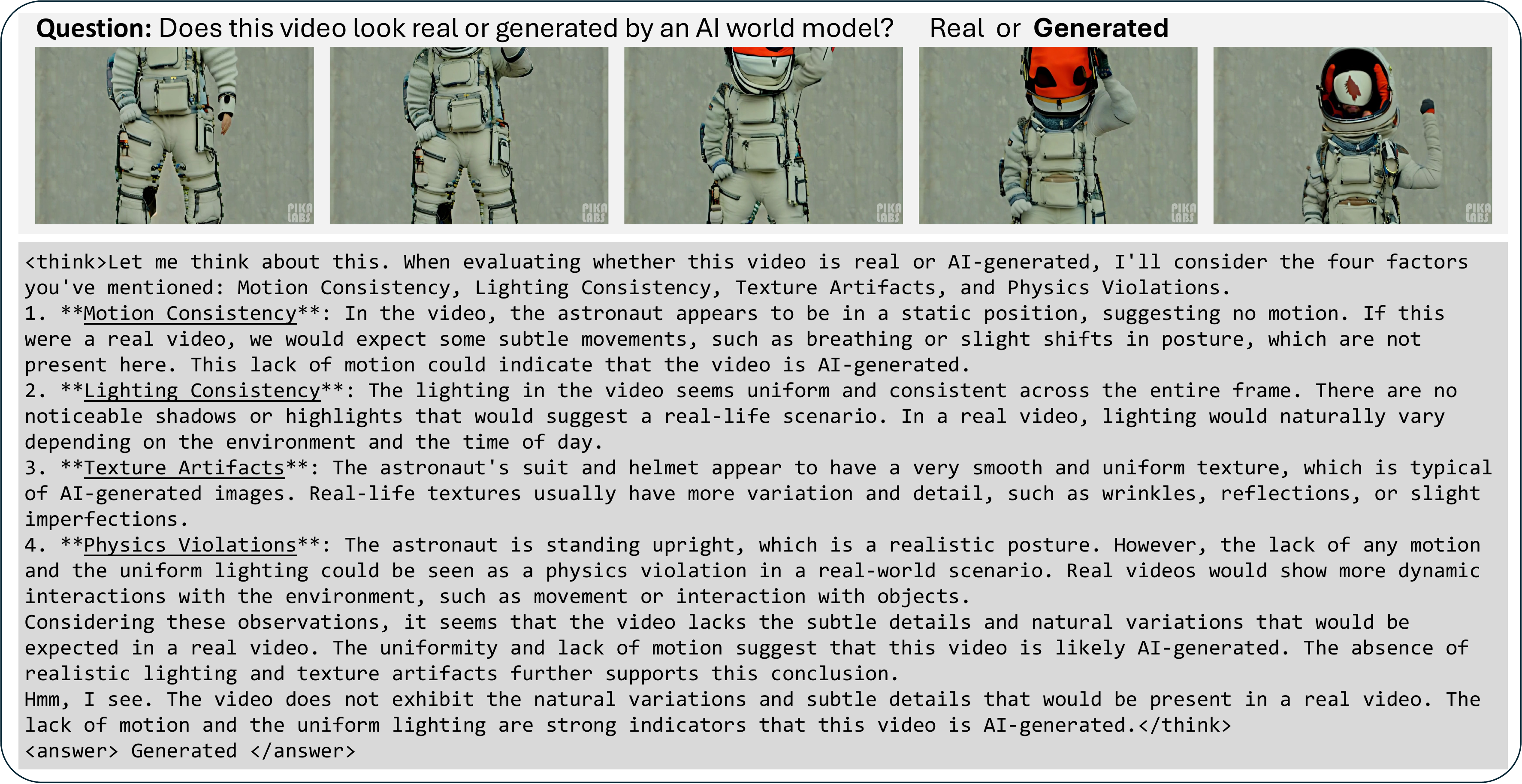}
    \caption{VidGuard-R1 (GRPO) inference on a ModelScope-generated video}
   \label{fig:ex_genvideo_modelscope}
\end{figure*}

\begin{figure*}[!ht]
    \centering
    \includegraphics[width=1\textwidth]{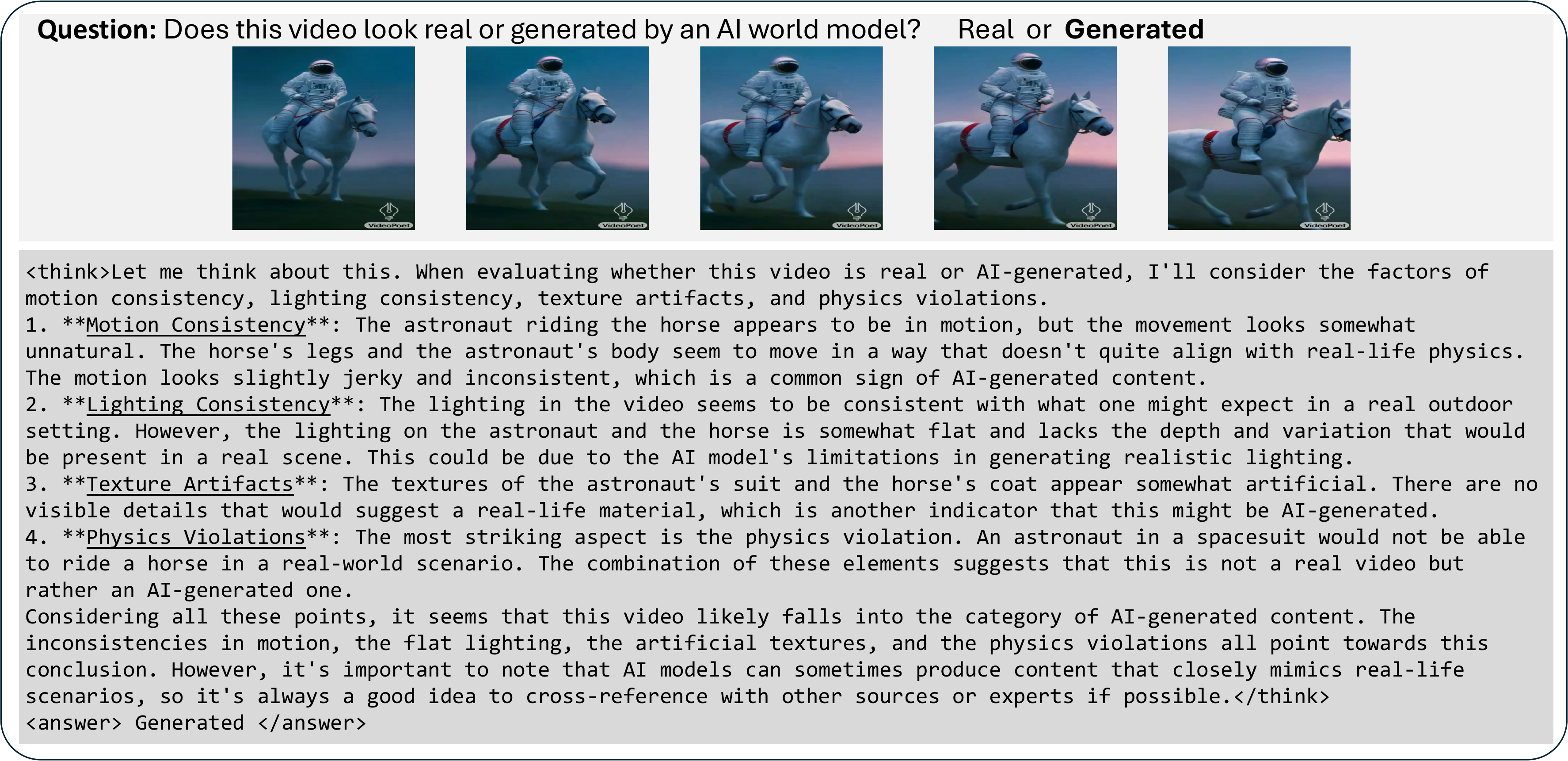}
    \caption{VidGuard-R1 (GRPO) inference on a DreamVideo-generated video}
   \label{fig:ex_genvideo_wildscrape}
\end{figure*}

Figures~\ref{fig:ex_genvideo_sora}--\ref{fig:ex_genvideo_wildscrape} show inference examples for videos generated by ten different AI models included in the GenVideo testing dataset: Sora~\citep{sora2025}, Morph Studio~\citep{morph}, Gen2~\citep{gen2}, HotShot~\citep{hotshot}, Lavie~\citep{lavie}, Show-1~\citep{show1}, Moonvalley~\citep{moonvalley}, Crafter~\citep{crafter}, ModelScope~\citep{modelscope}, and DreamVideo~\citep{dreamvideo}.
% sora, morph, gen2, hotshot, lavie, show-1, moon valley, crafter, modelscope, wildscrape

% musev, svd, cogvideo, mora


\clearpage

\end{document}